\documentclass{article}

\usepackage[final]{neurips_2022}

\usepackage[utf8]{inputenc} % allow utf-8 input
\usepackage[T1]{fontenc}    % use 8-bit T1 fonts
\usepackage{hyperref}       % hyperlinks
\usepackage{url}            % simple URL typesetting
\usepackage{booktabs}       % professional-quality tables
\usepackage{amsfonts}       % blackboard math symbols
\usepackage{nicefrac}       % compact symbols for 1/2, etc.
\usepackage{microtype}      % microtypography
\usepackage{xcolor}         % colors
\usepackage{times}
\usepackage{latexsym}
\usepackage{svg}
\usepackage{subcaption}
\usepackage{amssymb}

\usepackage[T1]{fontenc}
\usepackage[utf8]{inputenc}

\usepackage{array}
\usepackage{makecell}
\usepackage{multicol}

\usepackage{microtype}
\usepackage{amsfonts} 
\usepackage{algorithm2e}
\usepackage{algpseudocode}

\title{\textcolor{orange}{C}l\textcolor{orange}{a}ss\textcolor{orange}{if}iers are Better Experts for Controllable Text Generation}

\author{Askhat Sitdikov\thanks{$\;$ Equal contribution.} $\;$, Nikita Balagansky\footnotemark[1] $\;$, Daniil Gavrilov and Alexander Markov
       \\
       Tinkoff\\
           \{ext.asitdikov, n.n.balaganskiy, d.gavrilov, al.v.markov\}@tinkoff.ai
       }

\begin{document}
\maketitle
\begin{abstract}
This paper proposes a simple method for controllable text generation based on weighting logits with a free-form classifier, namely CAIF sampling. Using an arbitrary text classifier, we adjust a small part of a language model's logits and guide text generation towards or away from classifier prediction. 

We experimented with toxicity avoidance and sentiment control tasks and showed that the proposed method significantly outperforms recent PPLM, GeDi, and DExperts on PPL and task accuracy metrics based on the external classifier of generated texts. In addition, compared to other approaches, it is easier to implement and tune and has significantly fewer restrictions and requirements.
\end{abstract}

\section{Introduction}
\label{introduction}
Neural text generation is an important part of many NLP pipelines, such as those for dialog generation and question answering. However, the application of these models can be difficult without control over a Language Model (LM). For example, in order to apply a natural dialogue generation system, the model must not produce toxic or harmful texts.

One common way to control an LM is to guide its sampling process using a classifier to sample texts with desired properties (e.g., reduced toxicity). While PPLM \citep{pplm} uses an arbitrary text classifier to control an LM, most recent methods \citep{gedi, dexperts} rely on a classifier induced by LM conditioned on a certain topic \citet{ctrl}.

In this paper, we propose \textbf{C}l\textbf{a}ss\textbf{if}ier guided sampling (CAIF): a simple method for controllable text generation based on Bayesian re-weighting of LM logits using an external classifier. Unlike GeDi and DExperts, CAIF relies on a free-form classifier while being significantly easier to apply than PPLM.

There are two reasons why we consider using a free-form classifier for guiding LMs fascinating. 

The trivial reason is that this approach allows us to more easily perform controllable generation, since finding an existing arbitrary text classifier is much easier than a conditional LM. While a wide range of publicly available classifiers could be used to guide an LM, finding a conditional LM could be more challenging. When this paper was written, if one searches for a text classifier model with Hugging Face, then approximately $10$ thousand model could be accessed. Though no $10$k topics are available, and some of these models could even be trained on the same datasets, it still provides alternatives to guide the model. At the same time, only $23$ conditional language models could be found with search prompt \textit{topic}. Even if some other search prompts could provide some extra conditional language models, these rough statistics represent the situation with the availability of text classifiers and conditional LMs.

\newpage

\begin{figure}
 \centering
  \includegraphics[width=\columnwidth]{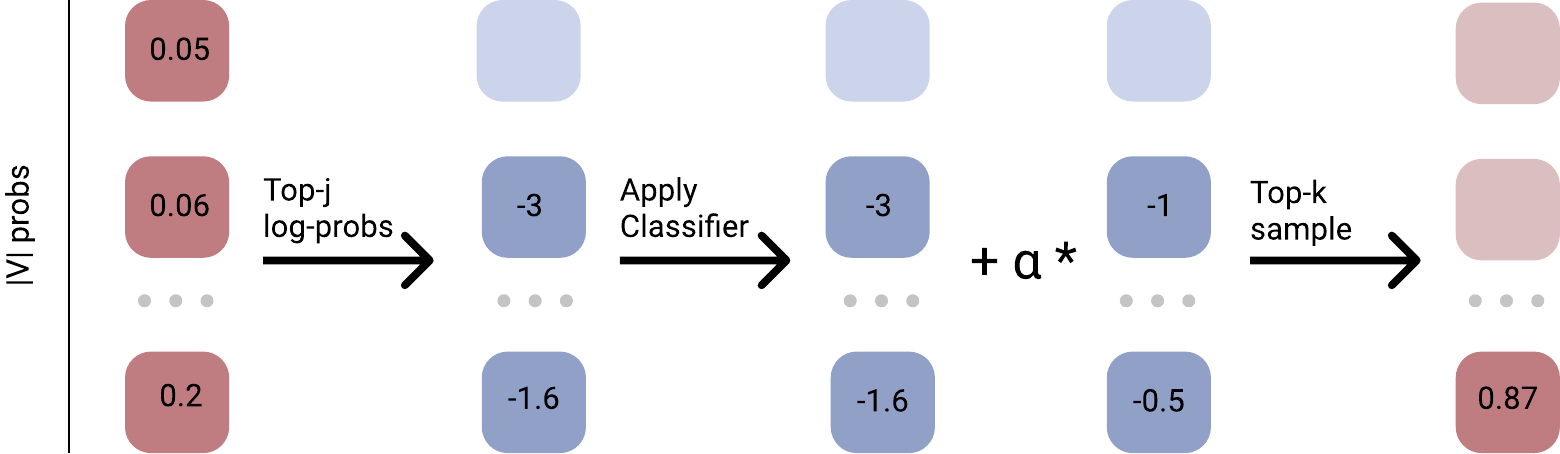}
  \caption{A schematic view of CAIF sampling. Having a probability distribution on tokens (with the total number of tokens equal to the size of vocabulary $|V|$), we select top-$j$ tokens to apply a classifier. We then add the logarithms of probabilities obtained from the classifier weighted by $\alpha \in \mathbb{R}$ to the logarithms of token probabilities and then select the top-$k$ tokens in order to sample the next token. Note that $k < j << |V|$.}
  \label{architecture-fig}
\end{figure}

The second reason can be considered more controversial. The wide adoption and success of both GeDi and DExperts \citep{gedi, dexperts} could make one assume that it is related to induced classifiers $\hat{p}(c|x)$ being capable of generalizing better due to their dependency on a smaller language model. With CAIF sampling we are answering the question of whether it is really necessary to apply a smaller LM to perform conditional text generation, or if it is enough to use a free-form classifier.

We experimented with the proposed approach and found that this simple method allowed us to significantly outperform recent detoxification approaches measured by perplexity (PPL) and sentiment accuracy. In addition, we explored the periodicity of applying a guiding mechanism. We showed that, while all recent methods guide an LM at each step, it is possible to guide a model depending on the entropy of an LM outputs' distribution. In order to get further insights into CAIF's limits, we explored its hyperparameters and showed that the range of the sampling weight hyperparameter could be extended to $\mathbb{R}$,  while previous works only used positive weight values.

% \section{Recent works}
% \citet{ctrl} proposed to train an LM on conditioned data, so one could control the generation by selecting a condition (CTRL). \citep{pplm} proposed PPLM, which uses an external classifier as a target for optimization of hidden states during the inference process (PPLM). 

% \begin{figure}
%   \centering

%   \medskip
%   \begin{subfigure}[t]{.49\linewidth}
%     \centering\includegraphics[width=\linewidth]{images/log_probs-2.pdf}
%     \caption{}
%   \end{subfigure}
%   \begin{subfigure}[t]{.49\linewidth}
%     \centering\includegraphics[width=\linewidth]{images/inv_probs.pdf}
%     \caption{}
%   \end{subfigure}

%   \caption{(a) A comparision of $-\log(x)$ and $\log(1 - x)$ scores which could be used for detoxification with classifier producing the toxicity probability $x$. For this plot, we used a fixed value of $\alpha = 1$. Note that $-\log(x)$ reduces quickly and assigns relatively low scores for $x > 0.2$, while $\log(1 - x)$ remains almost unchanged for $x < 0.4$. (b) A comparison of negative $\alpha$ with inverse probability sampling mechanisms. See Section \ref{alpha-selection} for more details.}
%   \label{logs}
% \end{figure}

% GeDi \citep{gedi} used an external LM with the desired topic or intent to use this LM as a classifier to perform importance sampling on next token probabilities. \citet{dexperts} proposed DExperts: a sampling mechanism based on the usage of two extra LMs conditioned towards and against the desired topic, which is used to reweight the probabilities of the next tokens.

\section{Recent Works}
\citet{ctrl} proposed to train an LM on conditioned data, so that generation could be controlled by selecting a condition (CTRL). However, it is important to consider that such a mechanism would require re-training the whole model in order to add new guiding topics.

\citet{pplm} proposed PPLM, which uses an external classifier as a target for optimization of hidden states during the inference process. While PPLM seems easy to implement, it hides a huge amount of haziness in the details. E.g., it is unclear whether it is necessary to optimize all hidden states in temporary dimensions.

GeDi \citep{gedi} used an external LM with the desired topic or intent as a classifier for re-weighting next token probabilities. \citet{dexperts} proposed DExperts, a sampling mechanism based on using two extra LMs conditioned towards and against the desired topic, which is used to reweight the probabilities of the next tokens. We argue that these methods can also be considered impractical. DExperts requires two additional LMs that are conditioned on positive and negative sentiments to perform controllable sampling, and GeDi uses an external conditioned LM as a classifier to perform re-weighting of LM logits. 
% We argue that dependency on external LMs is impractical. Training LMs could be difficult and require large amounts of data while training a stand-alone classifier is significantly easier.

\citet{fudge} explored the usage of a text classifier for controllable text generation with FUDGE. Although, FUDGE omitted the usage of hyperparameters used to control the strength of guiding (which we observed to be crucial for the conditional generation with CAIF in later sections). Furthermore, \citet{fudge} had not used a \textit{free-form} classifier since they explored the use of classifiers trained to perform inference with truncated sequences that occur during text generation.

\begin{figure*}[ht!]
  \centering
  \fontsize{12}{14}\selectfont

    \includegraphics[width=\linewidth]{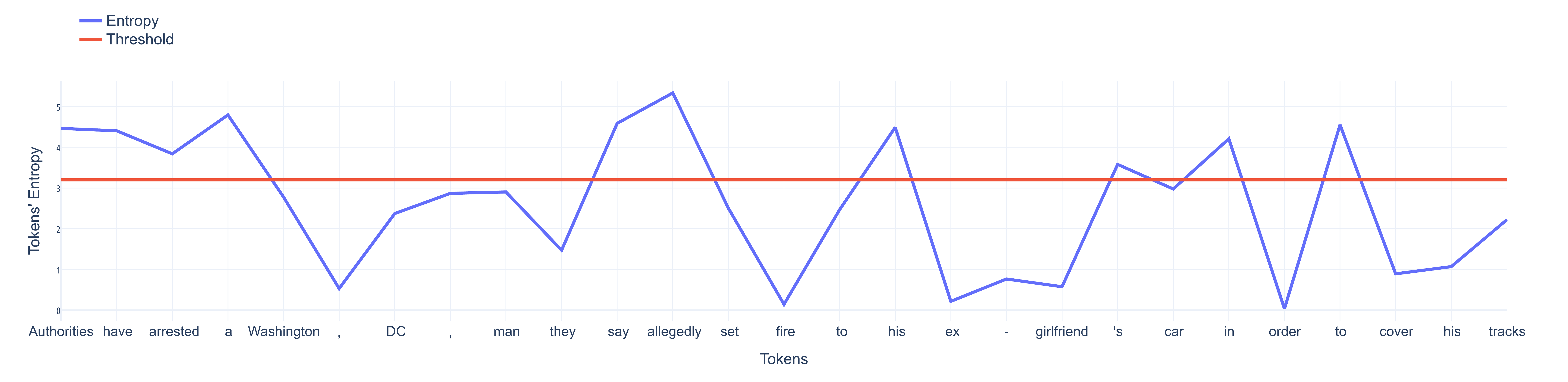}

  \caption{An example of entropy values across different tokens for a text prompt. Empirically, tokens at positions where LM outputs with low entropy, have a small impact on the semantics of text. Thus, one could perform a step of controllable text generation only for such LM outputs, in which entropy is larger than a threshold value. See Section \ref{entropy-section} for details.}
  \label{entropy-example}
\end{figure*}

\section{Background}
\label{background}
% A usual LM used to sample unconditioned text is based on autoregressive factorization of the probability of some text $x$. E.g., if $x$ is tokenized as a sequence $x_1, x_2, ..., x_n$, then the probability is modelled as

% \begin{equation}
% p(x|\theta) = \prod_i^n p(x_i|x_{<i}, \theta),
% \end{equation}

% where $\theta$ are parameters of a model optimized by maximizing the likelihood of training data (we will omit them later for simplicity), once a model is trained, one could sample a new text token-by-token (i.e., sample $x_1$ first, then sample $x_2$ and so on).

% While one could sample from $p(x_i|x_{<i})$ at each step, such generation strategy leads to gibberish samplings \citep{topp}. The most common ways to improve the quality of samplings are sampling with temperature \citep{temperature}, top-$k$ sampling \citep{topk}, and nucleus (top-$p$) sampling \citep{topp}.

% \subsection{Controllable text generation}
Controllable text generation could be seen as modeling a conditional text probability:

\begin{equation}
p(x|c) = \prod_i^n p(x_i|x_{<i}, c),
\end{equation}

where $c$ is an arbitrary condition (e.g., a topic or intent). If there is enough data for each necessary condition, training such a model from scratch is trivial. However, if that is not the case, training a well-performing LM may become difficult. A possible solution to this problem is inference-time controllable generation, which aims to adjust an unconditional $p(x)$ towards a conditional $p(x|c)$.

The most straightforward solution for inference-time control over an LM is re-weighting its logits using Bayesian inference in order to obtain a conditional $p(x_i|x_{<i}, c)$ out of unconditional $p(x_i|x_{<i})$ and an arbitrary classifier $p(c|x)$, as follows:

\begin{equation}
p(x_i|x_{<i}, c) \propto p(x_i|x_{<i})p(c|x_{\leq i})^{\alpha},
\end{equation}

where $\alpha$ is a hyperparameter modifying the importance of the classifier during sampling.

Sampling from such a model requires applying the classifier $p(c|x_{\leq i})$ during sampling at each step for each new possible token. In general cases, this significantly reduces the speed of this method's naive application. 

In order to overcome the problem of speed, \citet{gedi} proposed to use a conditioned LM. In their method, a small conditional LM $\hat{p}(x_i|x_{<i}, c)$ is inverted using Bayesian inference to obtain $\hat{p}(c|x_{\leq i})$, which is induced from an LM classifier and produces classification probabilities for all tokens at one step. Furthermore, it is possible to cache hidden states of $\hat{p}(x_i|x_{<i}, c)$ during sampling to increase inference speed even further.

However, as we noted in Section \ref{introduction}, dependency on an external conditional LM $\hat{p}(x|c)$ could be too harsh of a requirement to follow in practice. Thus, we propose exploring using free-form classifiers to guide an LM.

% The only reason to use $\hat{p}(x|c)$ is that it allows us to get all classification probabilities at once, instead of applying the classifier to each token in vocabulary (which could be large enough, e.g., tens of thousands of tokens).

\section{CAIF Sampling}
\label{caif}

\subsection{Proposed Method}
We argue that guiding an LM with a classifier $\hat{p}(c|x)$ induced from a smaller conditional LM $\hat{p}(x|c)$ is mostly done to improve inference speed. Thus, if we want to perform a controllable text generation with a free-form classifier, it is necessary to improve generation speed.

As we noted in Section \ref{background}, the main complexity of applying an arbitrary classifier is inevitable to evaluate class probability $p(c|x_{\leq i}) \triangleq p(c|x_{<i}, x_i)$ for each possible token $x_i$ at $i$-th position if we want to evaluate full $p(x_i|x_{<i}, c)$. Because the vocabulary size $|V|$ could easily reach tens of thousands of tokens, such a task would require an enormous amount of computations to sample a sequence.

This paper proposes simplifying re-weighting probabilities for controllable text generation by truncating the set of classified tokens. This idea is based on the observation that, while it is necessary to evaluate $p(c|x_{\leq i})$ for each token in vocabulary, sampling strategies (e.g., top-k sampling) will truncate most tokens with the lowest probabilities $p(x_i|x_{<i}, c)$. Therefore, some tokens with low probability $p(x_i|x_{<i})$ are not going to be considered for sampling even after weighting with large $p(c|x_{\leq i})$, and thus can be omitted from classification. 

Based on this heuristic, we propose CAIF sampling. During the sampling procedure, we use only $j$ tokens with the highest probability of being the next token $p(x_i|x_{<i})$ to evaluate a classifier on. Then, these top-$j$ tokens are reweighted and used for top-$k$ sampling. Note that here, $k < j << |V|$. See Figure \ref{architecture-fig} for a schematic view of the proposed algorithm. We observed that $j$ could be considered small, as it does not exceed $100$ tokens for classification during our experiments.

\subsection{Further Speeding Up CAIF}

While reducing the number of classified sequences during the sampling procedure dramatically improves inference speed, one could go even further. We could choose to re-weight LM logits for some specific steps instead of the entirety of the generation process. In the following subsections, we will describe possible approaches to doing so.

\subsubsection{Periodic Criterion for CAIF}
While the straightforward way of performing CAIF sampling is to apply a classifier at each generation step, it is possible to alternate CAIF sampling with plain sampling. 

More formally, we define CAIF sampling with period-$p$ as a generation strategy, where we adjust token probabilities at every $p$-th step. From this perspective, plain CAIF sampling could be seen as sampling with a period-$1$.

However, such a criterion could be seen as too harsh. There is no clear intuition behind applying CAIF periodically. Even if we were to sample a sequence with period-$2$ and guide the generation towards non-toxic texts, a model could still produce a toxic token at every $2$-nd step when CAIF is not applied.

\subsubsection{Entropy Criterion for CAIF}
\label{entropy-section}
\citet{entropy} hypothesized that the entropy of token probabilities represents the importance of the next token in the text. More concretely, if the entropy is low, then the next token in the sequence has a utilitarian role and vice versa. See Figure \ref{entropy-example} for an example of entropy values produced by GPT-2 Large for a text prompt.

From such a perspective, we could define CAIF sampling with entropy-$e$, where $e$ is a threshold entropy value. For this method, we only apply CAIF at such steps if the prediction entropy is greater than the threshold value.

Note that, similar to Periodic CAIF, plain CAIF can be seen as CAIF with entropy-$0$.

\section{Experiments}

\subsection{Experimental Setup}
% \subsubsection{Datasets}

% We followed the experimental setup of \citet{dexperts} in our experiments and used 10k non-toxic prompts from the RealToxicityPrompts dataset, alongside 5k neutral prompts and 2.5k negative prompts from OpenWebText Corpus.

\subsubsection{Toxicity Avoidance with RealToxicityPrompts Dataset}
\label{toxicity-section}

We followed the experimental setup of \citet{dexperts} in our experiments and used 10k non-toxic prompts from the RealToxicityPrompts dataset \citep{realtoxicity} to evaluate the ability of the proposed method to avoid toxicity in samples.

We sampled $25$ continuations for $10$k non-toxic prompts and evaluated the samplings' PPL and its diversity as the number of distinct $n$-grams normalized by the length of generated sequences. 

\begin{figure*}[h!]
  \centering

  \medskip
    \begin{subfigure}[t]{.24\linewidth}
    \centering\includegraphics[width=\linewidth]{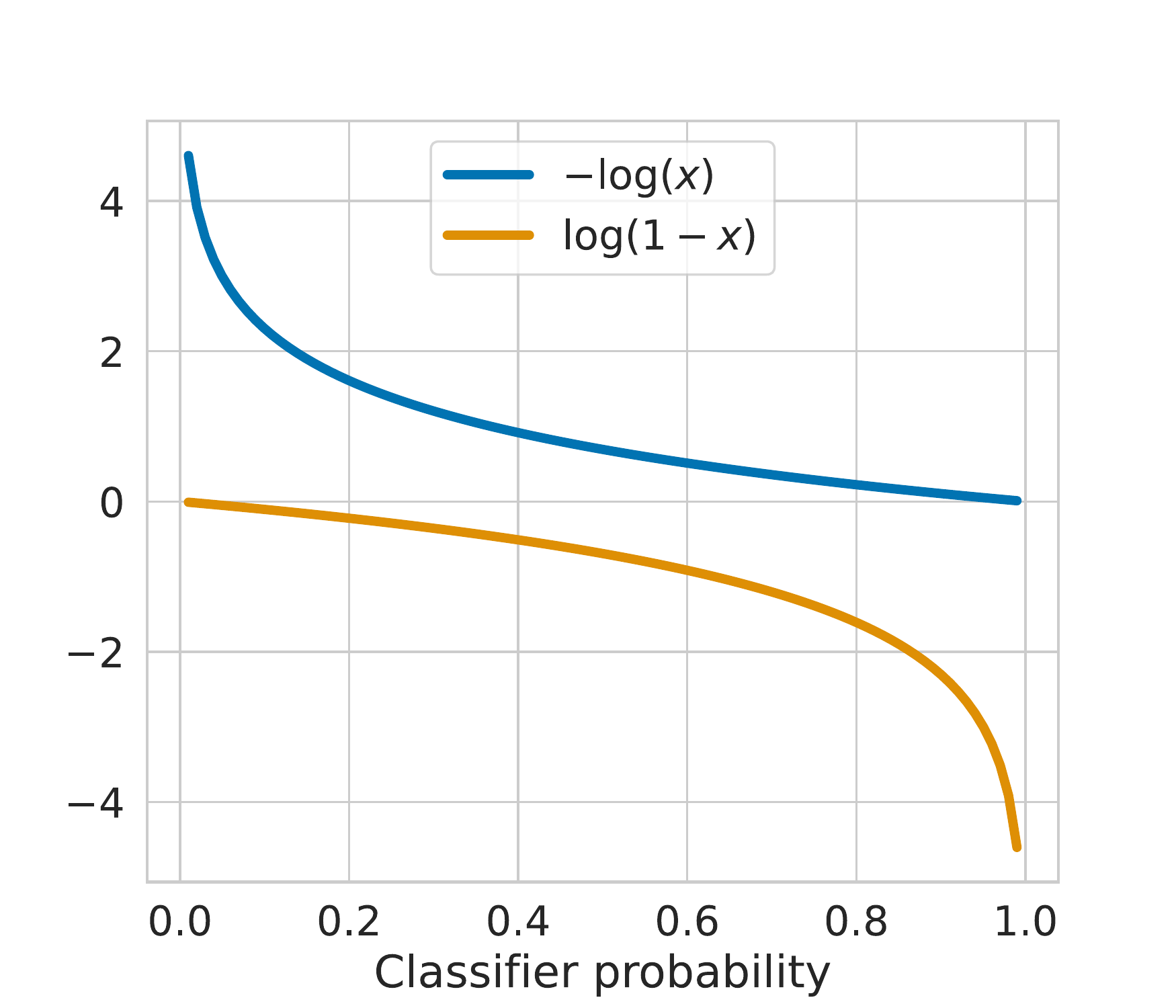}
    \caption{}
  \end{subfigure}
  \begin{subfigure}[t]{.24\linewidth}
    \centering\includegraphics[width=\linewidth]{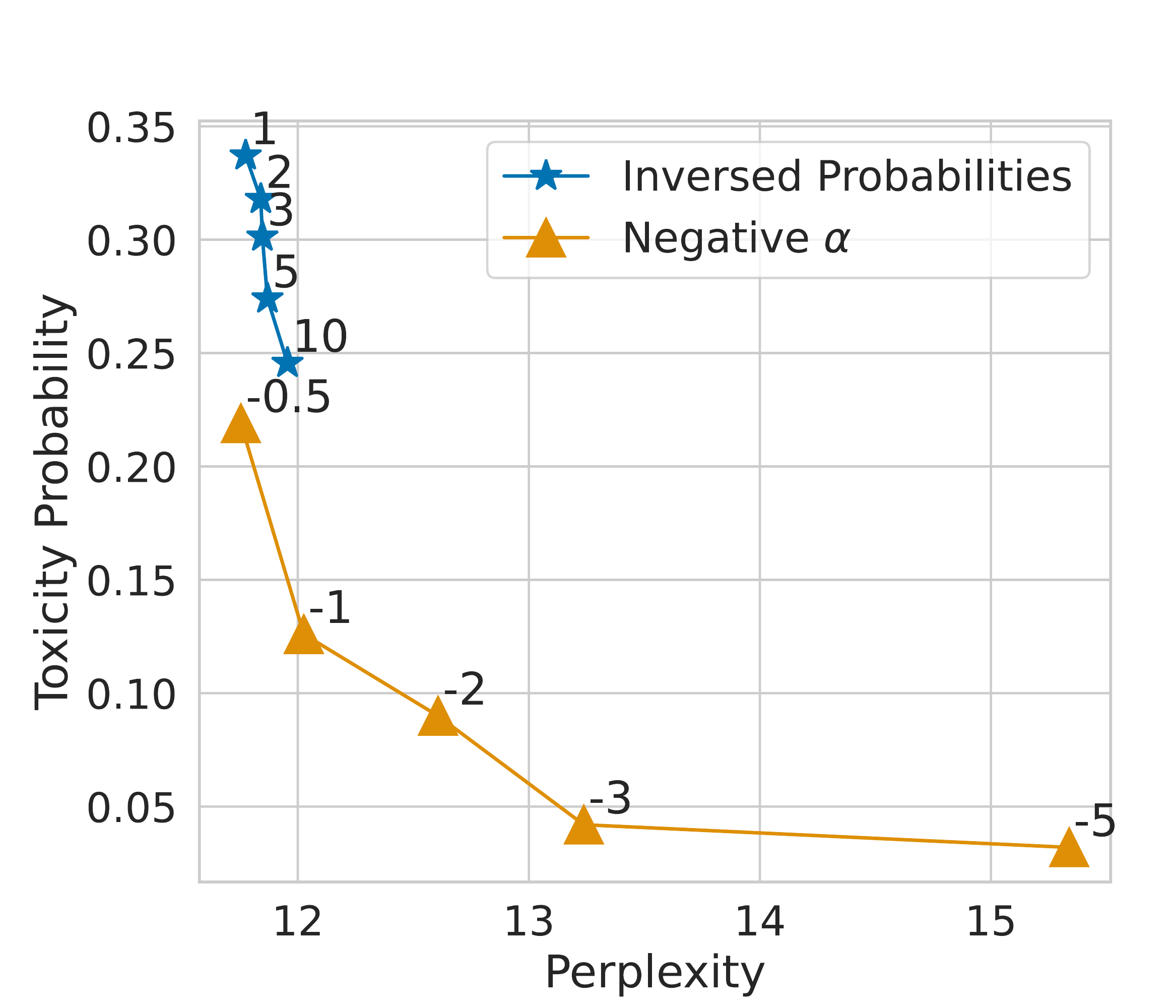}
    \caption{}
  \end{subfigure}
  \begin{subfigure}[t]{.24\linewidth}
    \centering\includegraphics[width=\linewidth]{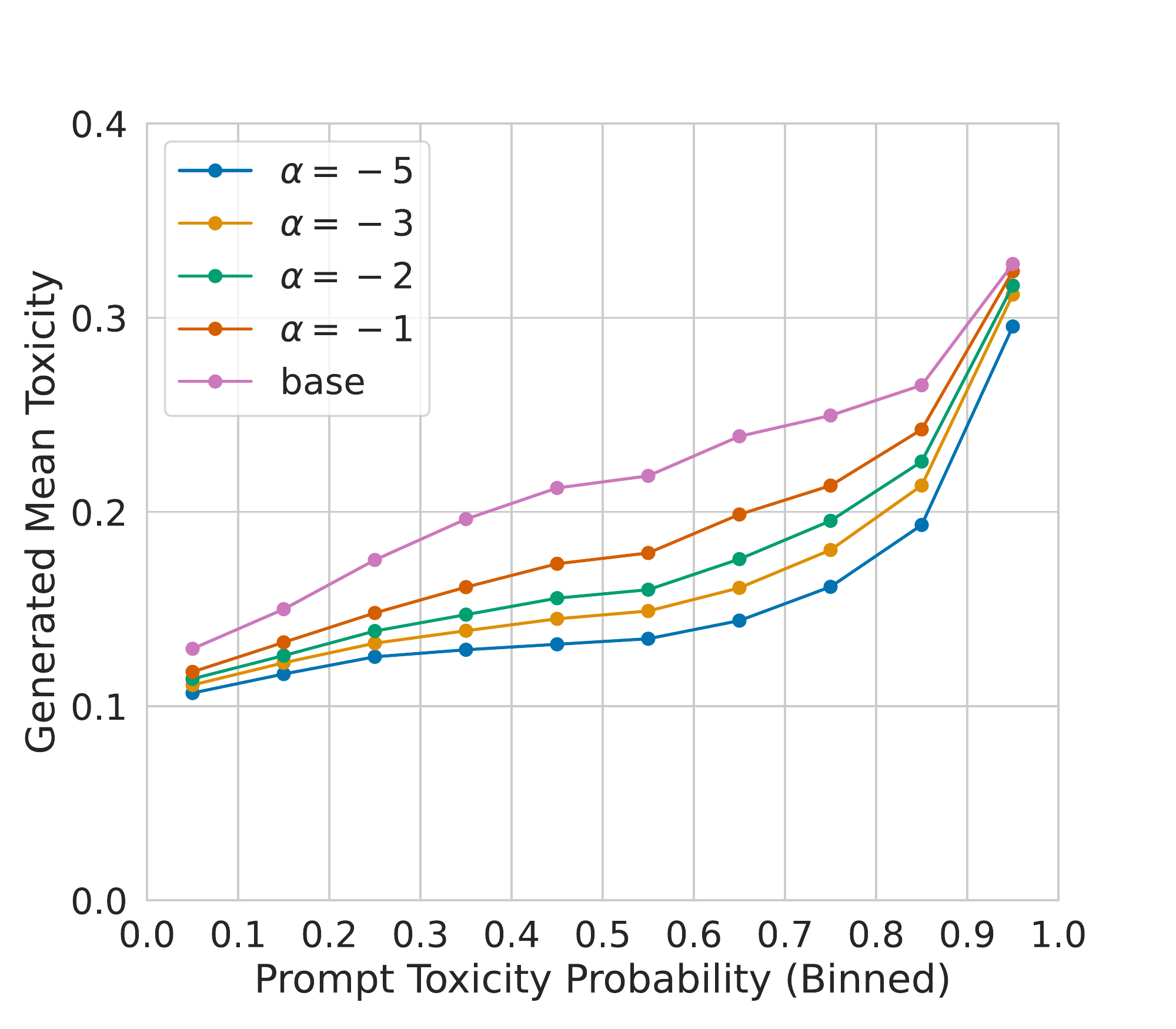}
    \caption{}
  \end{subfigure}
  \begin{subfigure}[t]{.24\linewidth}
    \centering\includegraphics[width=\linewidth]{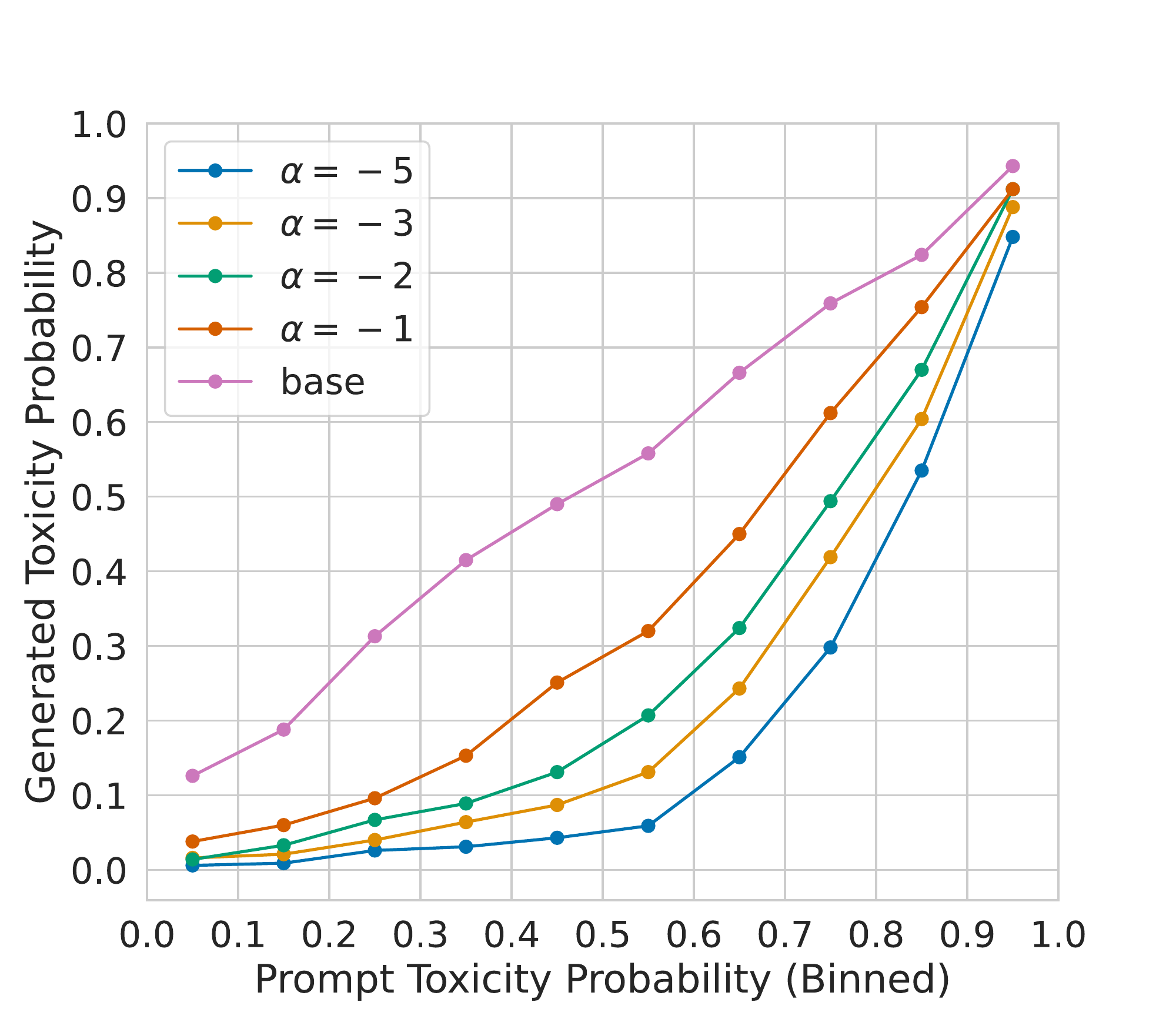}
    \caption{}
  \end{subfigure}

  \caption{(a) A comparison of $-\log(x)$ and $\log(1 - x)$ scores which could be used for detoxification with the classifier producing the toxicity probability $x$. For this plot, we used a fixed value of $\alpha = 1$. Note that $-\log(x)$ reduces quickly and assigns relatively low scores for $x > 0.2$, while $\log(1 - x)$ remains almost unchanged for $x < 0.4$. (b) A comparison of negative $\alpha$ with inverse probability sampling mechanisms. We report $\alpha$ values next to the plots. (c-d) A comparison of different $\alpha$ values with binned prompts from the RealToxicityPrompts Dataset with mean toxicity and toxicity probability metrics. See Section \ref{alpha-selection} for more details.}
  \label{bins}
\end{figure*}

We also evaluated the average mean and max toxicity level, alongside the empirical probability of occurrence of at least one negative sequence across $25$ samplings for each prompt. To evaluate the toxicity of generated sequences, we used the cardiffnlp/twitter-roberta-base-offensive classifier\footnote{https://huggingface.co/cardiffnlp/twitter-roberta-base-offensive} \citep{tweeteval}. To evaluate the perplexity of the sampled sequences, we used a pre-trained GPT-2 XL \citep{gpt2}. Following \citet{dexperts}, we evaluated the toxicity metric only for the generated part of sequences, omitting prompts.

As a base model for our experiments, we used GPT-2 Large, for which we applied different methods of controllable generation. For CAIF guiding we used the unitary/toxic-bert\footnote{https://huggingface.co/unitary/toxic-bert} classifier.

Also note that RealToxicityPrompts provides the labeling of toxicity levels for prompts in the dataset. With such labeling, we can divide the dataset into bins and evaluate baselines for each bin separately.

\subsubsection{Sentiment Control with OpenWebText Corpus}
\label{sentiment-section}
Following \citet{dexperts}, we used 5k neutral prompts and 2.5k negative prompts from OpenWebText Corpus\footnote{\citet{dexperts}, as well as experimented with positive prompts guided towards a negative sentiment. However, this experiment was omitted from this paper due to possible concerns regarding its practicality and ethics.}.

We used the mean percentage of positive samplings across all prompts as a metric for this experiment, as well as the PPL of samplings. To evaluate the positiveness of samplings, we applied distilbert-base-uncased-finetuned-sst-2-english\footnote{https://huggingface.co/distilbert-base-uncased-finetuned-sst-2-english} classifier. As for toxicity avoidance, we followed the setup from \citet{dexperts} and evaluated the sentiment of samplings on both prompts and continuations.

Following the experimental setup with Toxicity Avoidance (see Section \ref{toxicity-section}), we used GPT-2 Large as the model for generation and the cardiffnlp/twitter-roberta-base-sentiment\footnote{https://huggingface.co/cardiffnlp/twitter-roberta-base-sentiment} classifier \citep{tweeteval} to guide CAIF.

\subsection{Selection of $\alpha$}
\label{alpha-selection}

While \citet{gedi} only used $\alpha \geq 1$, we observed that we could use any $\alpha \in \mathbb{R}$. Suppose that we have a toxicity classifier which provides higher logit values as the input text increases in toxicity. In this case, the natural way to manage detoxification is to weight LM outputs at the $i$-th step with $\Big(1 - p(c|x_{\leq i})\Big)^{\alpha}$ and $\alpha > 0$ (namely, inverse probability weighting). However, we observed that its possible to perform weighting with $p(c|x_{\leq i})^{\alpha}$ and $\alpha < 0$ in order to reduce the toxicity of generated samples (negative $\alpha$).

Both $-\alpha\log(x)$ and $\alpha\log(1 - x)$ are decreasing functions on $x \in (0; 1)$ if $\alpha > 0$, which means that the highest score of importance sampling will be obtained when toxicity probability is at its lowest. However, a score obtained from a $-\alpha\log(x)$ dramatically reduces with a small increase of $x$, while $\alpha\log(1 - x)$ remains almost unchanged until a large value of $x$ is reached. See Figure \ref{bins}(a) for details.

We compared both of these detoxification approaches on the RealToxicityPrompts Dataset. We used CAIF sampling with a period-$1$ and top-$j = 100$ for both models and limited the dataset size to $1$k non-toxic prompts. See Figure \ref{bins}(b) for the comparison of negative $\alpha$ and inverse probability weighting. We observed that negative $\alpha$ showed a significantly better detoxification level while having better PPL values. As a result, instead of inverse probability, we used a negative $\alpha$ value in all further experiments for both toxicity avoidance and sentiment control tasks.

We also explored different values of $\alpha$. For this experiment, we used toxicity labels provided with the RealToxicityPrompt Dataset and evaluated CAIF using different toxicity levels of text prompts. See Figure \ref{bins} (c-d) for the results. We observed that CAIF, compared to plain sampling, drastically reduces the probability of toxic samples occurring, while higher absolute values of $\alpha$ avoid toxicity in samples better.

\begin{figure*}
  \centering

    \begin{subfigure}[t]{.24\linewidth}
      \includegraphics[width=\linewidth]{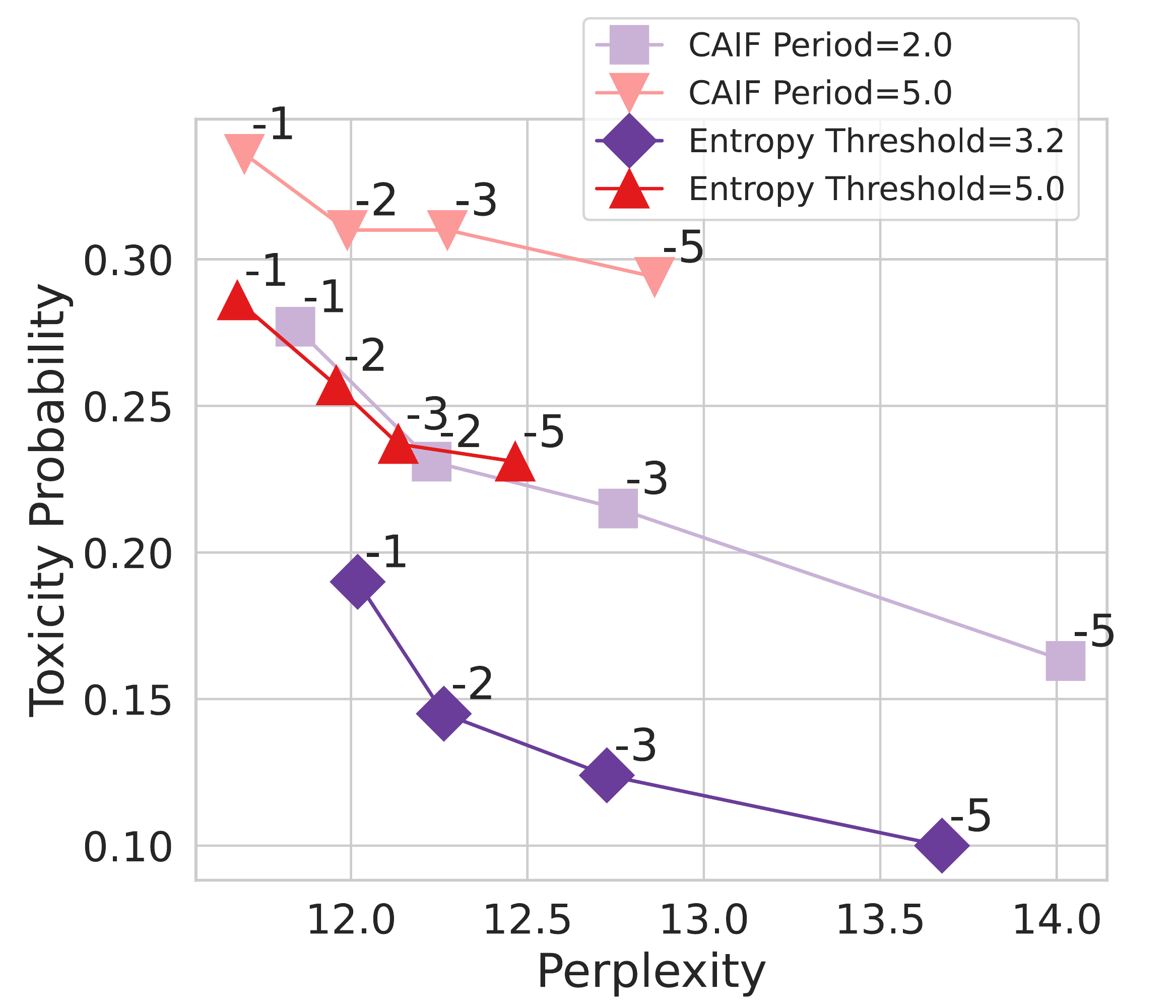}
      \caption{}
    \end{subfigure}
  \begin{subfigure}[t]{.24\linewidth}
    \centering\includegraphics[width=\linewidth]{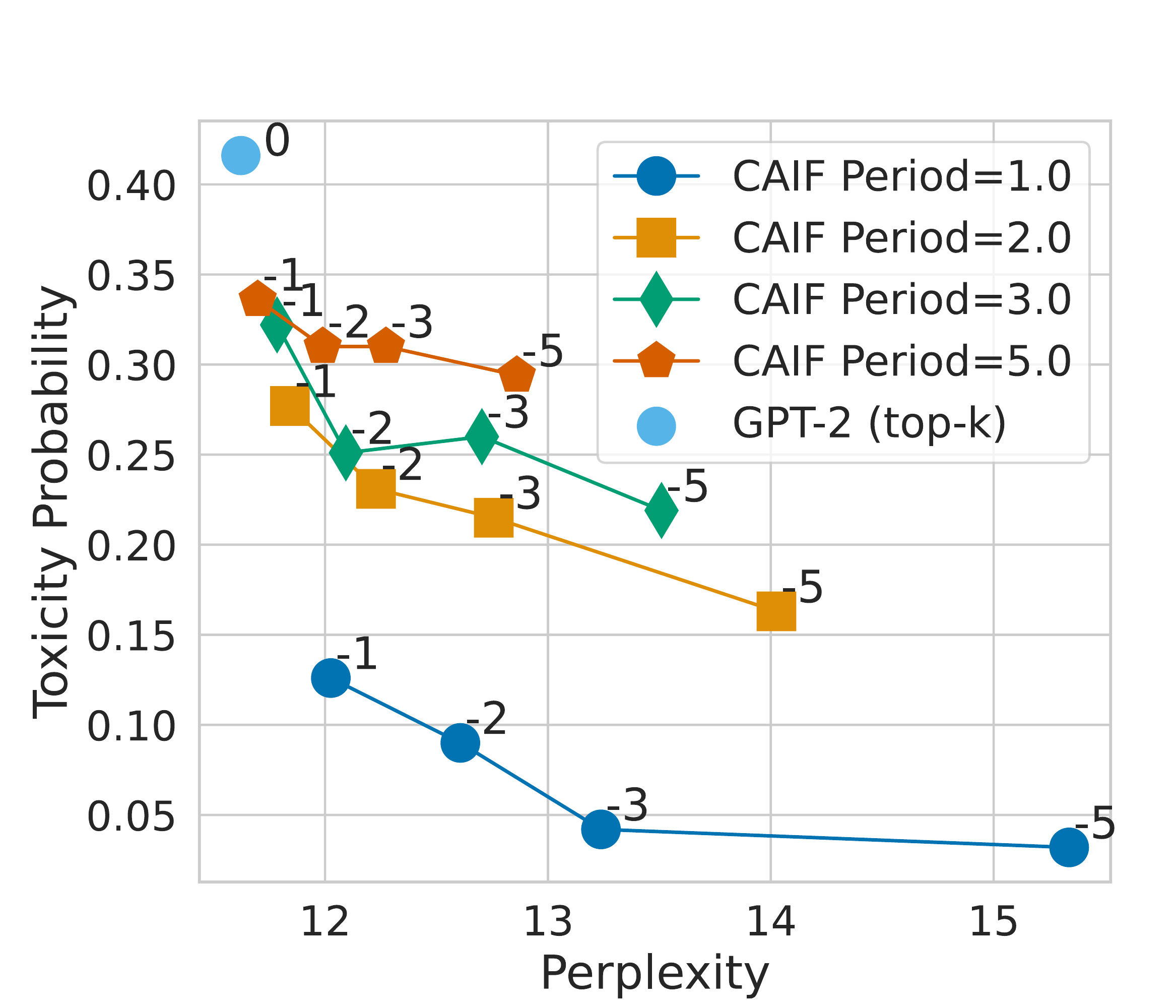}
    \caption{}
  \end{subfigure}
  \begin{subfigure}[t]{.24\linewidth}
    \centering\includegraphics[width=\linewidth]{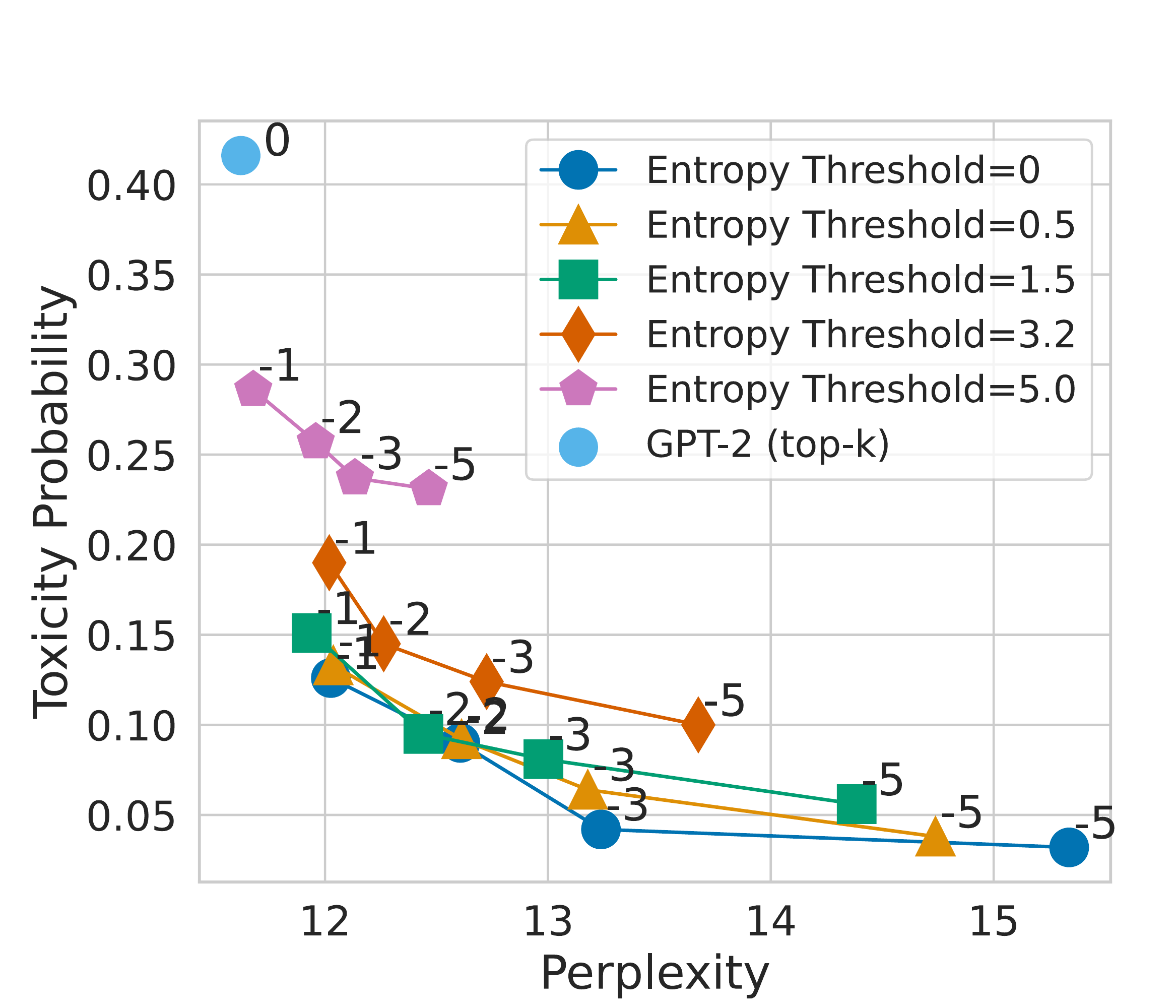}
    \caption{}
  \end{subfigure}
    \begin{subfigure}[t]{.24\linewidth}
    \centering\includegraphics[width=\linewidth]{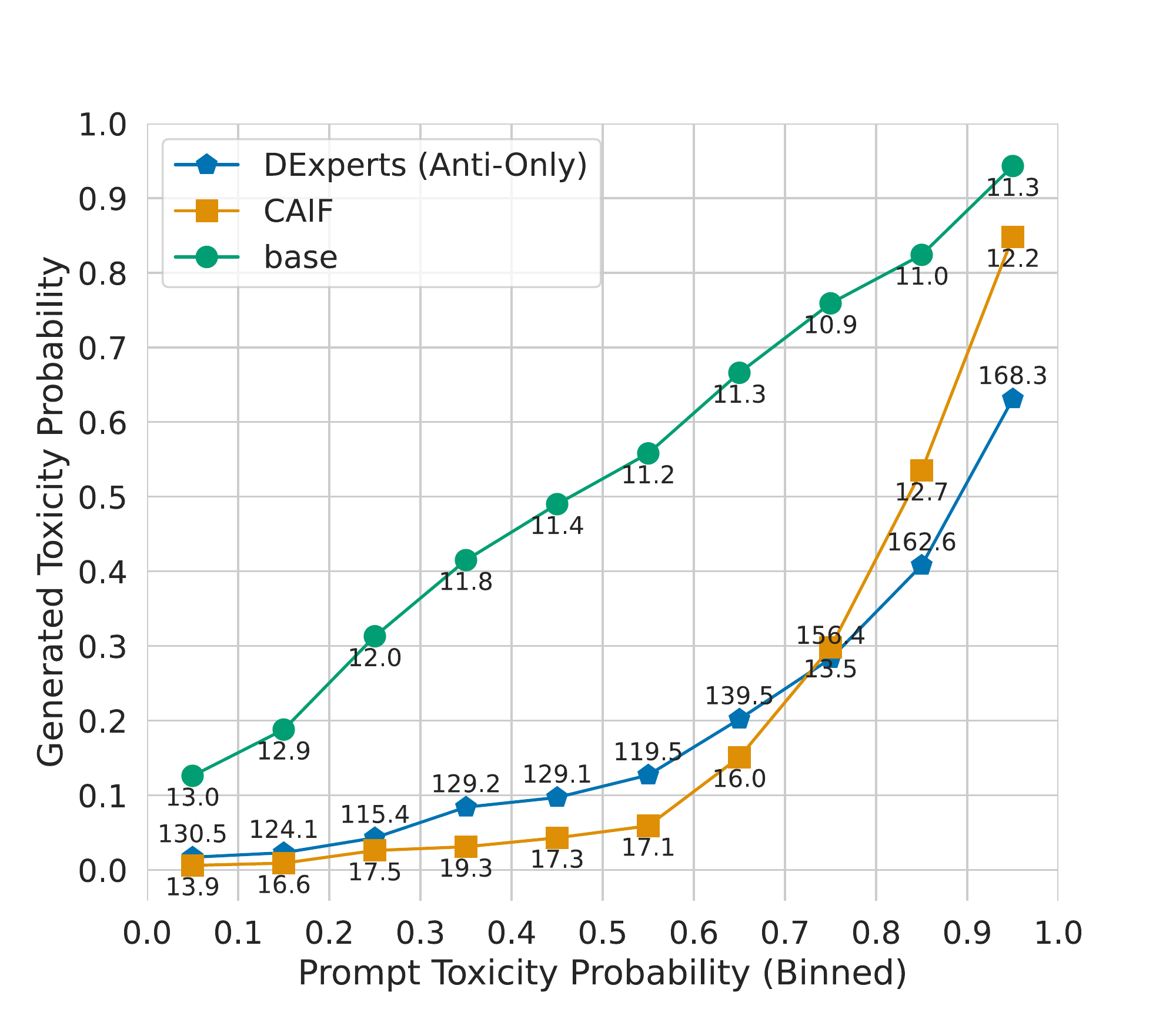}
    \caption{}
  \end{subfigure}

  \caption{(a) A comparison between periodic and entropy CAIF samplings with the comparable proportion of guided steps for the toxicity avoidance task. We report $\alpha$ values next to the plots. (b-c) A comparison of periodic CAIF with entropy CAIF samplings on the toxicity avoidance task. We report $\alpha$ values next to the plots. See more details in Section \ref{understanding-period-section}. (d) A comparison of CAIF sampling with DExperts with binned prompts from RealToxicityPrompts Datasets and toxicity probability metric. We also report PPL across samplings for each bin on the plots. See Section \ref{avoidance} for more details.}
  \label{threshold_vs_period}
\end{figure*}

\subsection{Understanding the Period of CAIF}
\label{understanding-period-section}

We compared plain CAIF sampling with Periodic CAIF and Entropy CAIF on the toxicity avoidance task with 1k prompts. For this experiment, we selected such periods $p$ and entropy thresholds $e$ that the proportion of guided steps would be comparable (see Section \ref{speed-section} for a detailed experiment with sampling speed). While the evaluation of this proportion is trivial for periodic CAIF (period-$2$ corresponds to 50$\%$ of guided tokens), for entropy CAIF, we evaluated empiric CDF of entropy across model outputs (see Appendix Figure \ref{entropy-cdf-images}). Based on this CDF, we could compare periods $2$ and $5$ with entropy of $3.2$ and $5.0$, which is 50$\%$ and 20$\%$ of guided steps compared to unguided ones. 
See Figure \ref{threshold_vs_period}(a) for the results. We observed that entropy CAIF performed marginally better than the periodic criterion measured by both PPL and toxicity probability metrics. See Figures \ref{threshold_vs_period}(b-c) to view a broader range of periods and entropy thresholds. For these, we observed that entropy CAIF could perform with negligible performance loss compared to plain CAIF on the toxicity avoidance task.

% As for the toxicity avoidance task (see Section \ref{understanding-period-section}), we observed that entropy CAIF performs with negligible loss in the performance compared to plain CAIF. E.g., period-$0.5$ we perform $80\%$ of guided steps while still outperforming other baselines which guide an LM at each step.

\begin{table*}[ht!]
\resizebox{\linewidth}{!}{%
\begin{tabular}{|l|c|c|c|c|c|c|c|}
\hline
Sampling & PPL $\downarrow$ & mean tox. $\downarrow$ & max tox. $\downarrow$ & tox. prob. $\downarrow$ & dist 1 $\uparrow$ & dist 2 $\uparrow$ & dist 3 $\uparrow$ \\ \hline \hline
% GPT2 T=1.5 & 22.4 & 17.7 & 45.3 & 38.8 & 55.6 & 86.4 & 86.6 \\ \hline
GPT-2 & 25.5 & 18.2 & 47.5 & 43.1 & 57.9 & 85.2 & 85.2 \\ \hline
PPLM & 32.6 & 17.7 & 45.9 & 40.0 & 58.4 & \textbf{85.5} & \textbf{85.5} \\ \hline
GeDi & 60.0 & 13.7 & 32.2 & 11.2 & \textbf{61.5} & 83.9 & 82.7 \\ \hline
DExperts & 32.4 & 13.9 & 29.7 & 7.5 & 58.0 & 84.0 & 84.1 \\ \hline
DExperts (top-k) & 20.2 & 13.3 & 27.9 & 6.4 & 52.9 & 80.4 & 82.5 \\ \hline
CAIF (our) & \textbf{15.0} & \textbf{12.0} & \textbf{26.1} & \textbf{3.3} & 51.5 & 81.2 & 84.1 \\ \hline
\end{tabular}
}
\caption{Results on the toxicity avoidance task for $10$k non-toxic prompts. See Section \ref{avoidance} for more details.}
\label{caif-table}
\end{table*}

\subsection{Toxicity Avoidance}
\label{avoidance}
We compared CAIF sampling with PPLM, GeDi, and DExperts approaches on the toxicity avoidance task, for which we guided models towards low toxicity values (see Section \ref{toxicity-section} for details of experimental setup).

For CAIF sampling, we used top-$k = 20$, top-$j = 100$, $\alpha = -5.0$. For other baselines, we used top-$p$ sampling with $p = 0.9$ \citep{topp}. We also experimented with top-$k = 20$ on DExperts for consistency of comparison with CAIF, which is designed to work with top-$k$ sampling.

See Table \ref{caif-table} for the results from non-toxic prompts, and Appendix Table \ref{samples-table} for the sample generations. We observed that CAIF performed dramatically better than other baselines. We obtained a significantly lower toxicity level on all metrics while having lower PPL than other baselines. Although CAIF showed slightly worse results on $n$-gram repetition metrics because top-$k$ sampling was used, the loss in repetition is not dramatic when taking into account the gain in perplexity and toxicity.

We also compared CAIF sampling to the DExperts method with binned prompts from RealToxicityPrompts Dataset (see Figure \ref{threshold_vs_period}(d)). We observed that CAIF outperformed DExperts for bins with toxicity $< 0.75$. While DExperts showed lower toxicity probability for more toxic prompts, it also dramatically increased the PPL of samplings for such bins.

% \begin{figure}
%   \centering

%   \medskip
%   \begin{subfigure}[t]{\linewidth}
%     \centering\includegraphics[width=\linewidth]{images/detox.pdf}
%     \caption{}
%   \end{subfigure}
%   \caption{A comparison of different periods of CAIF sampling on the toxicity avoidance task for 1k non-toxic prompts. See more details in Section \ref{period}.}
%   \label{results-images}
% \end{figure}

% \begin{figure}
%   \centering
%   \fontsize{4}{6}\selectfont

%     \includegraphics[width=.95\linewidth]{images/bins.pdf}

%   \caption{A comparison of CAIF sampling with DExperts with binned prompts from RealToxicityPrompts Datasets and toxicity probability metric. We also report PPL across samplings for each bin on the plots. See Section \ref{avoidance} for more details.}
%   \label{bins-dexperts}
% \end{figure}

\begin{figure*}[ht!]
  \centering

  \medskip
  \begin{subfigure}[t]{.32\linewidth}
    \centering\includegraphics[width=\linewidth]{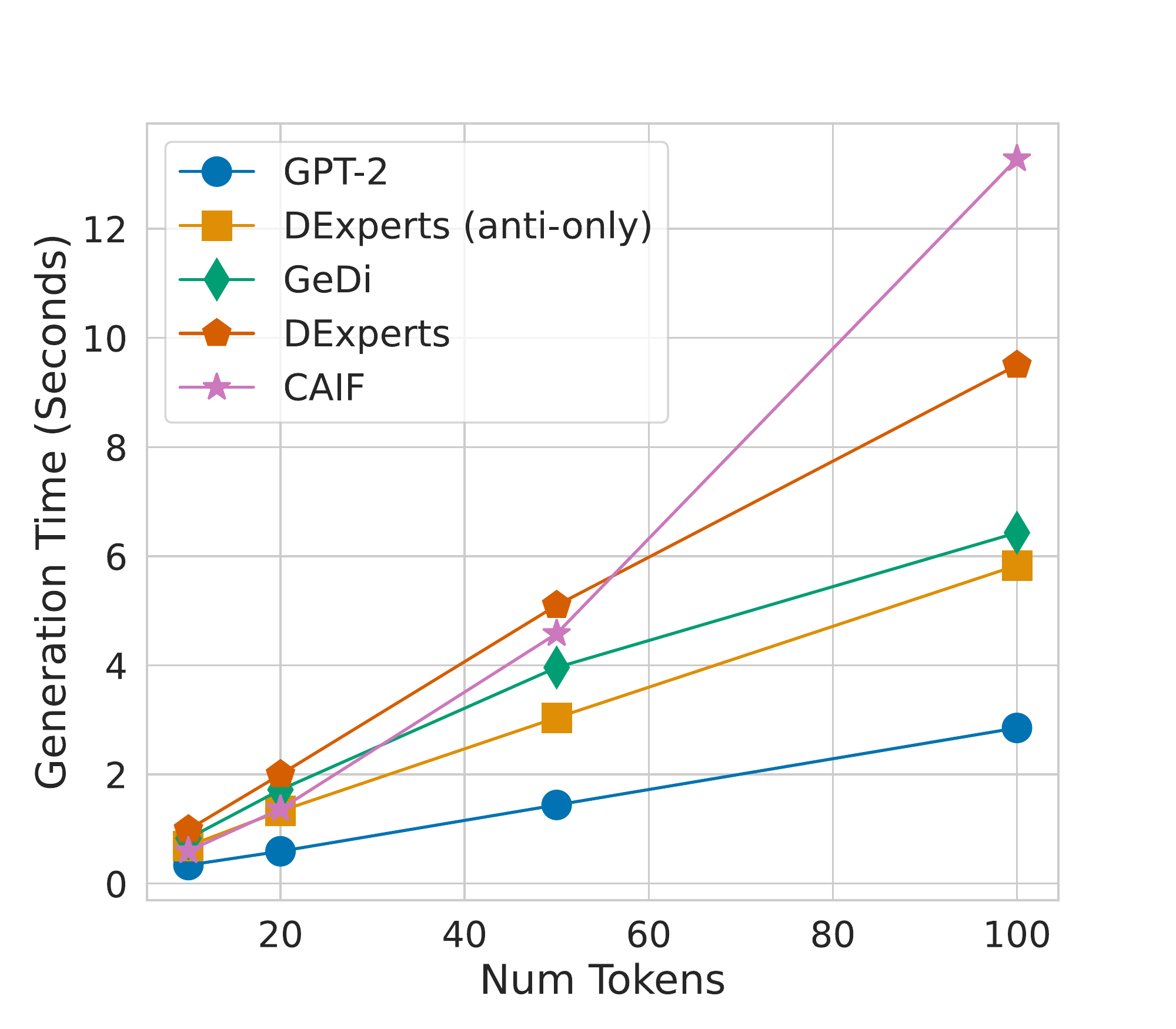}
    \caption{}
  \end{subfigure}
  \begin{subfigure}[t]{.32\linewidth}
    \centering\includegraphics[width=\linewidth]{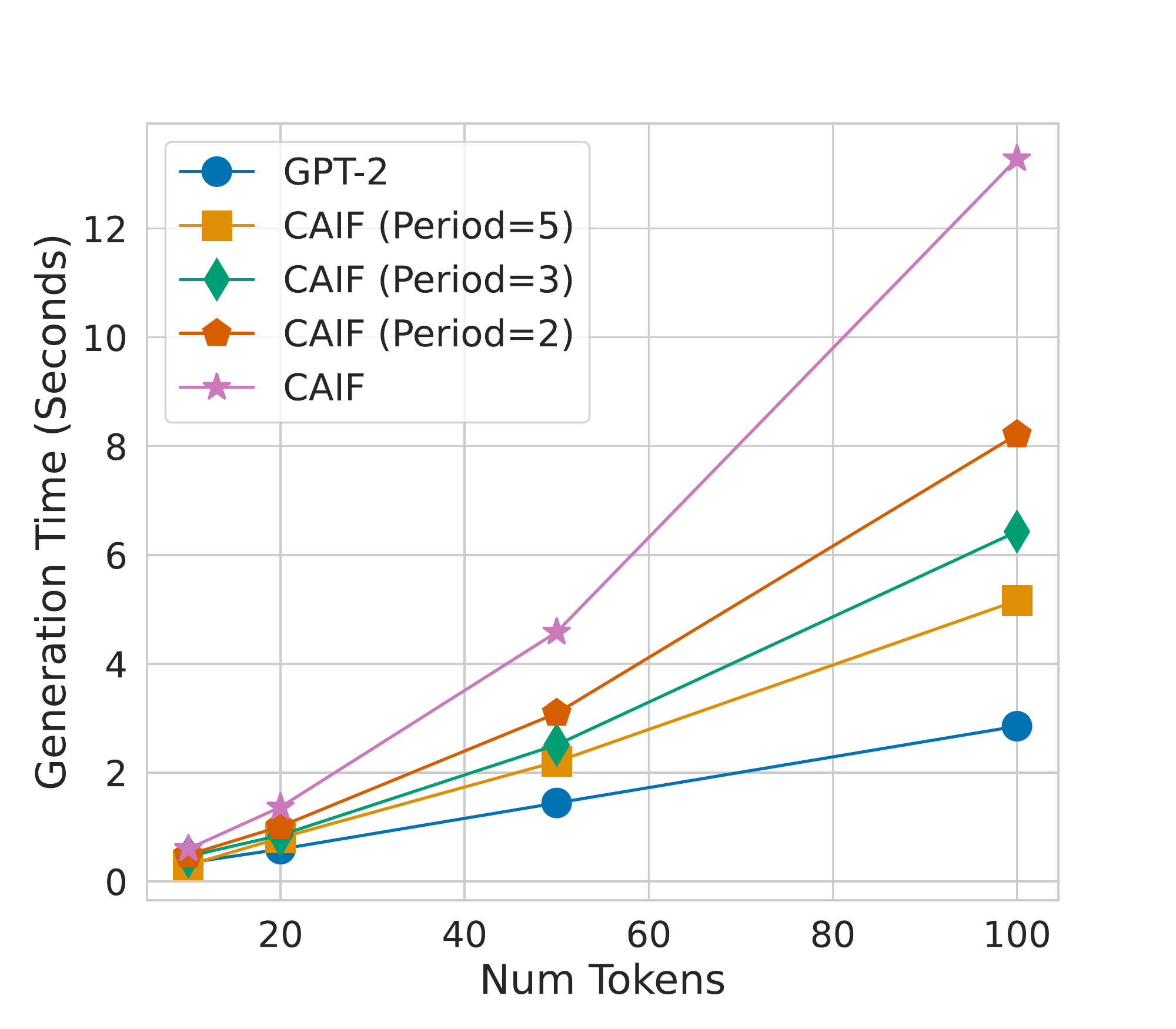}
    \caption{}
  \end{subfigure}
    \begin{subfigure}[t]{.32\linewidth}
    \centering\includegraphics[width=\linewidth]{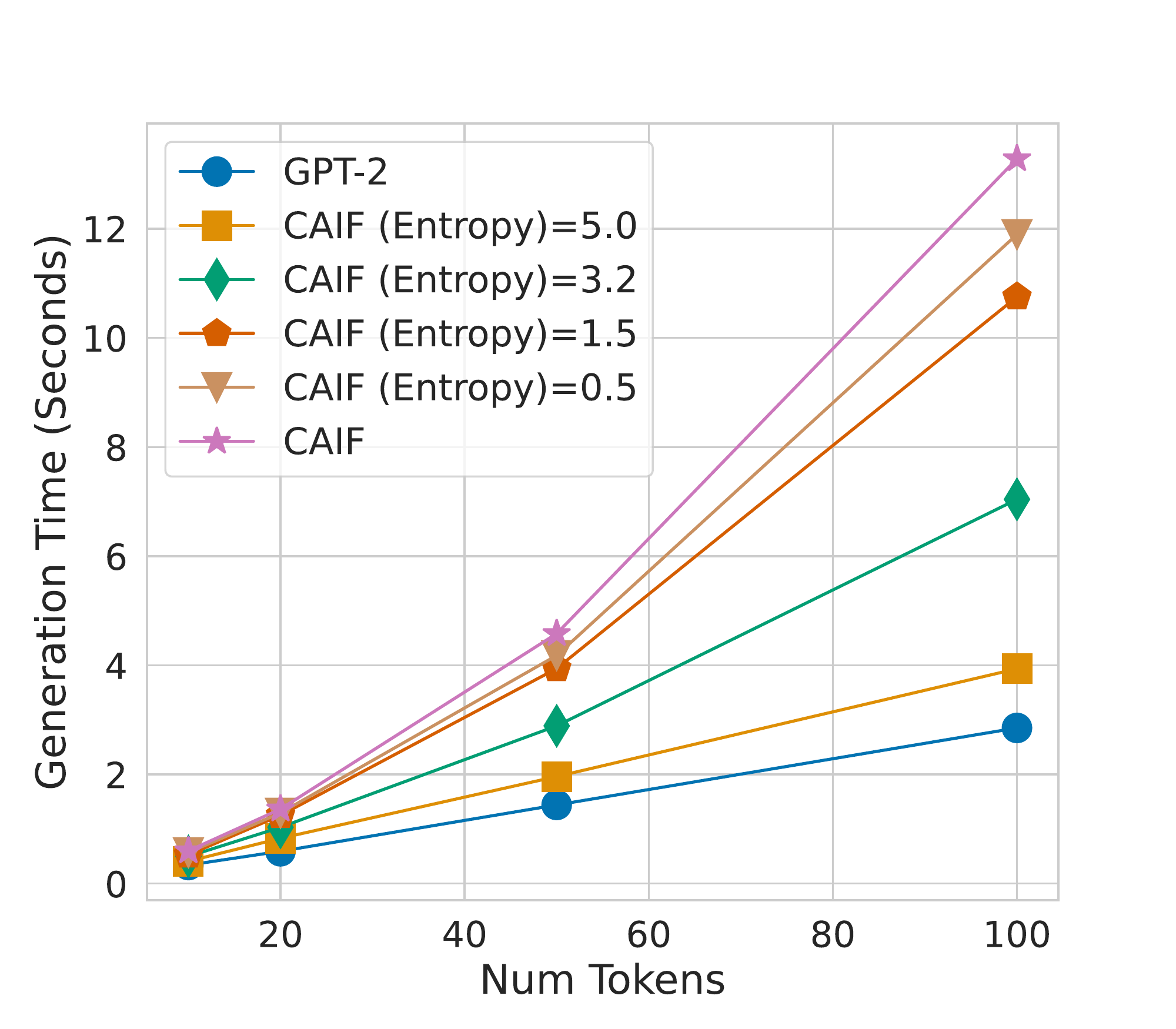}
    \caption{}
  \end{subfigure}

  \caption{Inference speed comparison of (a) CAIF and other related methods, among different CAIF periods (b) and entropy thresholds (c). }
  \label{speed}
\end{figure*}

\begin{figure}[ht!]
  \centering

  \medskip
%   \begin{subfigure}[t]{.33\linewidth}
%     \centering\includegraphics[width=\linewidth]{images/detox.pdf}
%     \caption{}
%   \end{subfigure}
  \begin{subfigure}[t]{.24\linewidth}
    \centering\includegraphics[width=\linewidth]{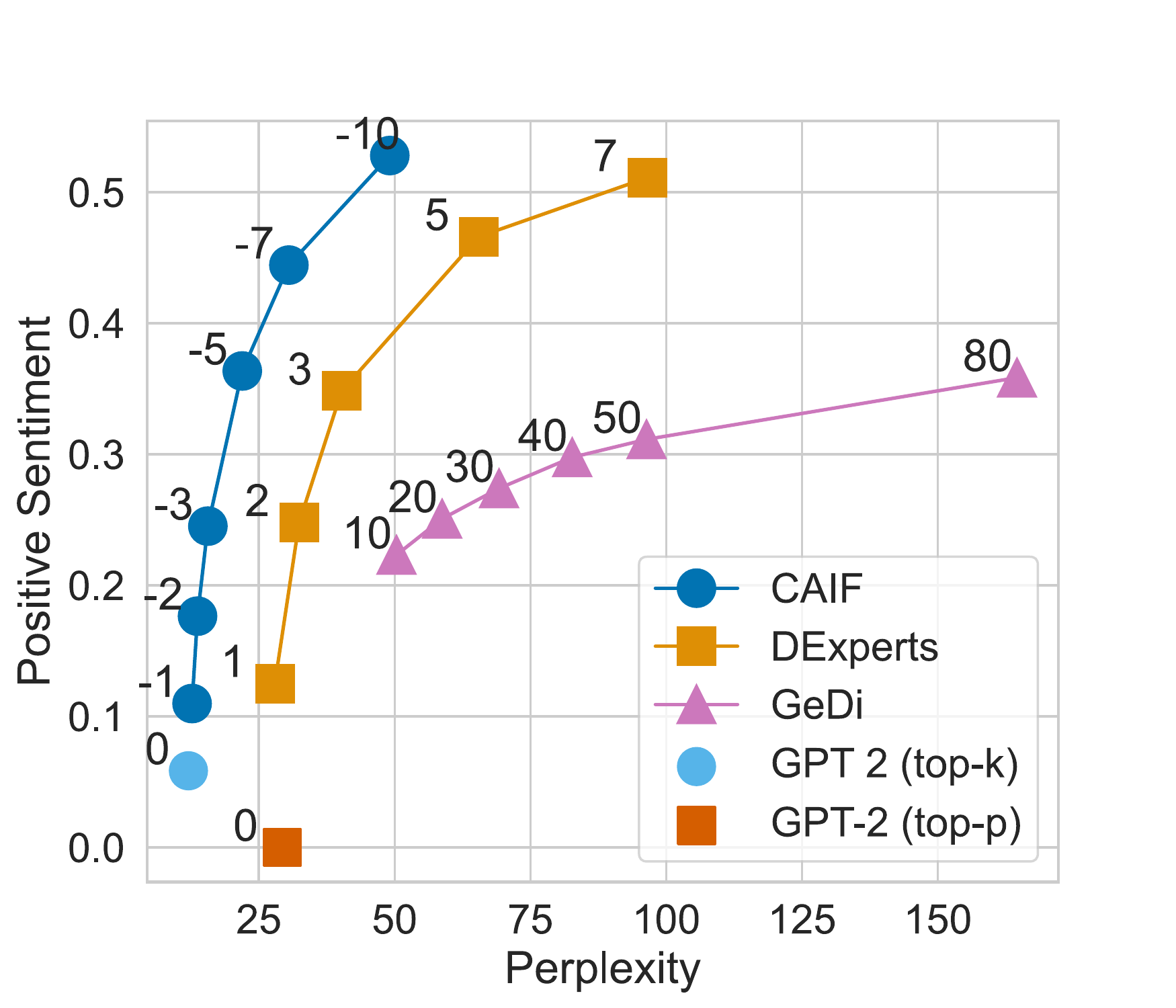}
    \caption{}
  \end{subfigure}
    \begin{subfigure}[t]{.24\linewidth}
    \centering\includegraphics[width=\linewidth]{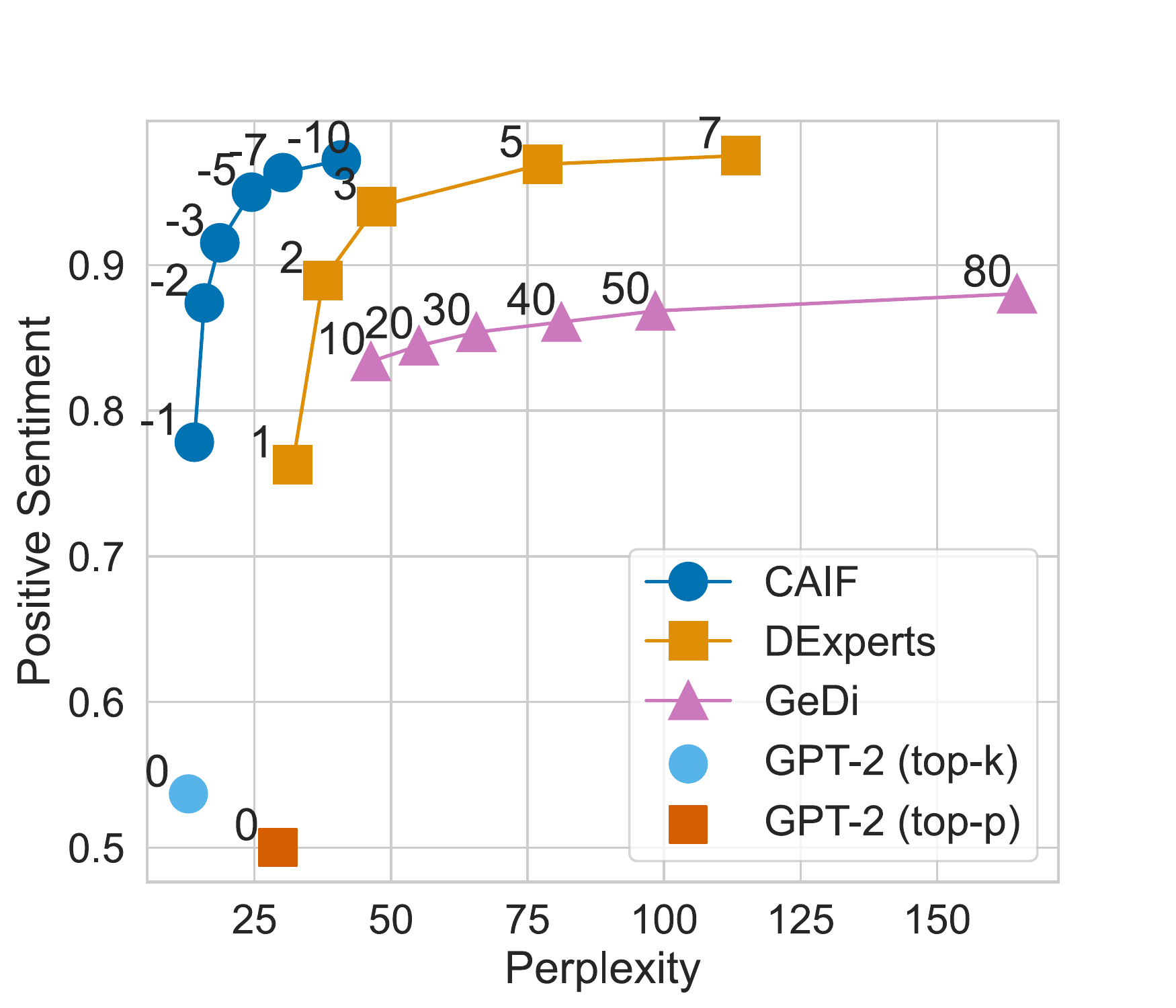}
    \caption{}
  \end{subfigure}
        \begin{subfigure}[t]{.24\linewidth}
    \centering\includegraphics[width=\linewidth]{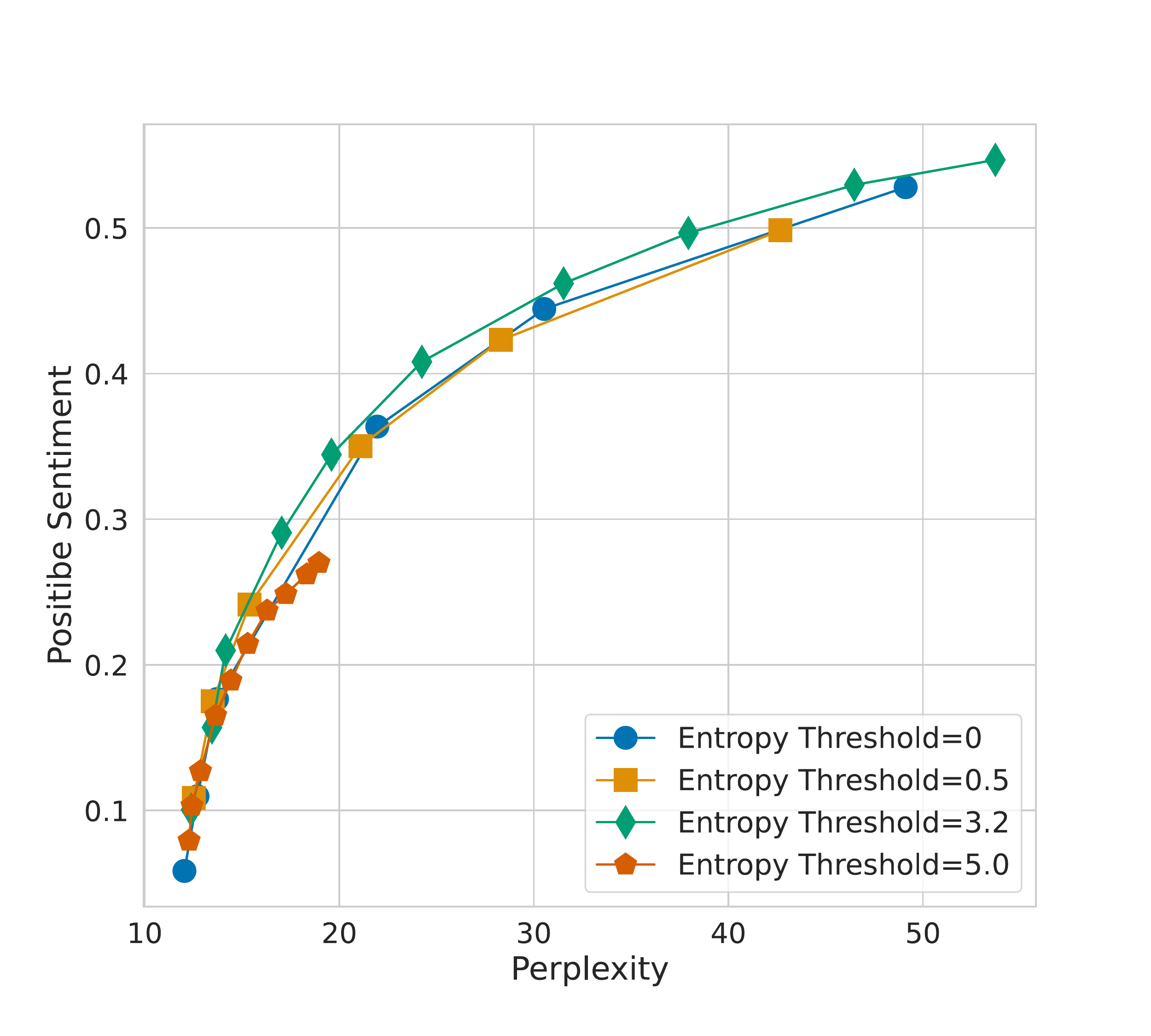}
    \caption{}
  \end{subfigure}
    \begin{subfigure}[t]{.24\linewidth}
    \centering\includegraphics[width=\linewidth]{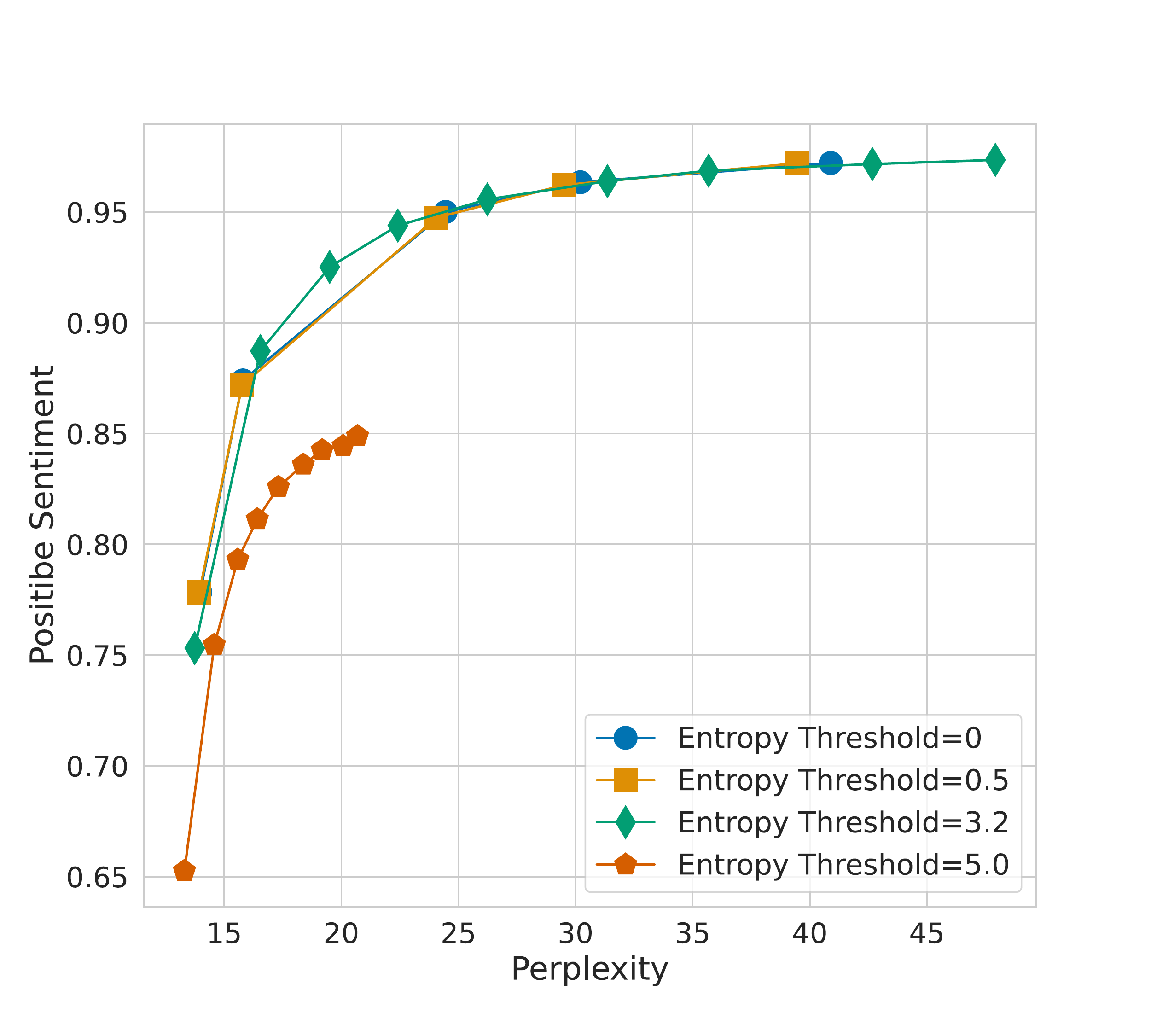}
    \caption{}
  \end{subfigure}

  \caption{(a-b) Sentiment control on negative and neutral prompt results. (c-d) A comparison of CAIF sampling with the Entropy CAIF criterion on sentiment control task for negative and neutral prompts. We omitted $\alpha$ values on these plots for visibility. Each dot represents $\alpha \in [-1, -2, -3, -5, -7, -10, -15, -20, -30, -40]$ for entropy thresholds $3.0$ and $5.0$, and $\alpha \in [-1, -2, -3, -5, -7, -10]$ for thresholds $0.5$ and $0.0$ from left to right. Threshold value $0$ represents plain CAIF. See section \ref{sentiment} for more details.}
  \label{results-images}
\end{figure}

\subsection{Sentiment Control}
\label{sentiment}
We compared CAIF with PPLM, GeDi, and DExperts on the sentiment control task (see Section \ref{sentiment-section} for details of the experimental setup).

See Figures \ref{results-images}(a-b) for the results. As for toxicity avoidance, CAIF performed dramatically better on both negative and neutral prompts, showing higher values of positiveness for samplings while having lower perplexity.

In addition, Figures \ref{results-images}(c-d) show a comparison of plain CAIF sampling with entropy CAIF on sentiment control. We observed that entropy CAIF, even with large entropy threshold values, performed comparable to plain CAIF (e.g, for neutral prompts, entropy-$3.2$ produced the same results as plain CAIF, even while using larger values of $\alpha$) or even outperformed it (for negative prompts, entropy-$3.2$ performed better than plain CAIF). These results are notable since only half of the performed steps were guided with a classifier for CAIF with entropy-$3.2$.

% \begin{figure}
%   \centering

%   \medskip
  
%       \begin{subfigure}[t]{.49\linewidth}
%     \centering\includegraphics[width=\linewidth]{images/negative.pdf}
%     \caption{}
%   \end{subfigure}
%     \begin{subfigure}[t]{.49\linewidth}
%     \centering\includegraphics[width=\linewidth]{images/neutral.pdf}
%     \caption{}
%   \end{subfigure}

%   \caption{A comparison of CAIF sampling with the Entropy CAIF criterion on sentiment control task for negative (a) and neutral (b) prompts. We omitted $\alpha$ values on these plots for visibility. Each dot represents $\alpha \in [-1, -2, -3, -5, -7, -10, -15, -20, -30, -40]$ for entropy thresholds $3.0$ and $5.0$, and $\alpha \in [-1, -2, -3, -5, -7, -10]$ for thresholds $0.5$ and $0.0$ from left to right. Threshold value $0$ represents plain CAIF. See section \ref{sentiment} for more details.}
%   \label{results-images-2}
% \end{figure}

\subsection{Sampling Speed}
\label{speed-section}
We evaluated the time necessary to sample a sequence with NVidia V100 GPU, a batch size equal to $1$ and sequence lengths in the range $n \in [10, 20, 50, 100]$. We compared CAIF with DExperts and GeDi approaches, for which we used the official implementation of evaluation. For CAIF, we used sampling with top-$j$ = 100 and top-$k$ = 20 \citep{topk}, while for DExperts, we used filter-$p$ = 0.9 and top-$k$ = 20. For each method, we report the mean value of wall-clock sampling time across $100$ runs.

See Figure \ref{speed}(a) for the results.  We observed that CAIF is comparable to other controllable generation methods in terms of speed for small sequence lengths (i.e., $n \leq 50$). For short sequences ($n \leq 20$), CAIF performed faster than other baselines. Note that with the growth of sequence length, CAIF requires more time to evaluate since using a free-form classifier requires $\mathbb{O}(n^2)$ time at each evaluation step. At the same time, GeDi and DExperts require only $\mathbb{O}(n)$ steps to evaluate thanks to caching used in induced LM classifiers. See Figures \ref{speed}(b-c) for the evaluation results of periodic and entropy CAIF samplings.

\section{Conclusion \& Future Work}
In this paper, we proposed a simple method of importance sampling approximation for controllable text generation. CAIF sampling showed dramatically better results than related approaches for toxicity avoidance and sentiment control tasks measured by PPL and task accuracy of samples.

We also performed a study of hyperparameters used in CAIF sampling and showed that weight $\alpha$ used for importance sampling could be drawn from $\mathbb{R}$ and not the previously used values of $\alpha \geq 1$.

In practical tasks (e.g., when a dialogue model is used), CAIF sampling is slower than other related methods, as several response candidates are generated and then filtered by a sentiment classifier to produce only positive responses. At the same time, a plug-and-play method for controllable generation allows us to develop a pipeline where no post-processing of samples is necessary, dramatically reducing the number of candidates that are necessary to sample. This shows the importance of PPL and toxicity level metrics of the method and the relative unimportance of sampling speed.

In this paper, we proposed two approaches for speeding up CAIF: Periodic and Entropy CAIF criteria, for which we alternate steps of plain sampling with guided sampling steps. We believe that CAIF could further benefit from new criterions of application.

\bibliography{custom}

\begin{thebibliography}{11}
\expandafter\ifx\csname natexlab\endcsname\relax\def\natexlab#1{#1}\fi

\bibitem[{Barbieri et~al.(2020)Barbieri, Camacho-Collados, Espinosa~Anke, and
  Neves}]{tweeteval}
Francesco Barbieri, Jose Camacho-Collados, Luis Espinosa~Anke, and Leonardo
  Neves. 2020.
\newblock \href {https://doi.org/10.18653/v1/2020.findings-emnlp.148}
  {{T}weet{E}val: Unified benchmark and comparative evaluation for tweet
  classification}.
\newblock In \emph{Findings of the Association for Computational Linguistics:
  EMNLP 2020}, pages 1644--1650, Online. Association for Computational
  Linguistics.

\bibitem[{Dathathri et~al.(2020)Dathathri, Madotto, Lan, Hung, Frank, Molino,
  Yosinski, and Liu}]{pplm}
Sumanth Dathathri, Andrea Madotto, Janice Lan, Jane Hung, Eric Frank, Piero
  Molino, Jason Yosinski, and Rosanne Liu. 2020.
\newblock \href {https://openreview.net/forum?id=H1edEyBKDS} {Plug and play
  language models: A simple approach to controlled text generation}.
\newblock In \emph{International Conference on Learning Representations}.

\bibitem[{Fan et~al.(2018)Fan, Lewis, and Dauphin}]{topk}
Angela Fan, Mike Lewis, and Yann Dauphin. 2018.
\newblock \href {https://doi.org/10.18653/v1/P18-1082} {Hierarchical neural
  story generation}.
\newblock pages 889--898.

\bibitem[{Gehman et~al.(2020)Gehman, Gururangan, Sap, Choi, and
  Smith}]{realtoxicity}
Samuel Gehman, Suchin Gururangan, Maarten Sap, Yejin Choi, and Noah~A. Smith.
  2020.
\newblock \href {https://doi.org/10.18653/v1/2020.findings-emnlp.301}
  {{R}eal{T}oxicity{P}rompts: Evaluating neural toxic degeneration in language
  models}.
\newblock In \emph{Findings of the Association for Computational Linguistics:
  EMNLP 2020}, pages 3356--3369, Online. Association for Computational
  Linguistics.

\bibitem[{Holtzman et~al.(2020)Holtzman, Buys, Du, Forbes, and Choi}]{topp}
Ari Holtzman, Jan Buys, Li~Du, Maxwell Forbes, and Yejin Choi. 2020.
\newblock \href {https://openreview.net/forum?id=rygGQyrFvH} {The curious case
  of neural text degeneration}.
\newblock In \emph{International Conference on Learning Representations}.

\bibitem[{Keskar et~al.(2019)Keskar, McCann, Varshney, Xiong, and
  Socher}]{ctrl}
Nitish~Shirish Keskar, Bryan McCann, Lav Varshney, Caiming Xiong, and Richard
  Socher. 2019.
\newblock {CTRL - A Conditional Transformer Language Model for Controllable
  Generation}.
\newblock \emph{arXiv preprint arXiv:1909.05858}.

\bibitem[{Krause et~al.(2020)Krause, Gotmare, McCann, Keskar, Joty, Socher, and
  Rajani}]{gedi}
Ben Krause, Akhilesh~Deepak Gotmare, Bryan McCann, Nitish~Shirish Keskar,
  Shafiq Joty, Richard Socher, and Nazneen~Fatema Rajani. 2020.
\newblock {GeDi: Generative Discriminator Guided Sequence Generation}.
\newblock \emph{arXiv preprint arXiv:2009.06367}.

\bibitem[{Liu et~al.(2021)Liu, Sap, Lu, Swayamdipta, Bhagavatula, Smith, and
  Choi}]{dexperts}
Alisa Liu, Maarten Sap, Ximing Lu, Swabha Swayamdipta, Chandra Bhagavatula,
  Noah~A. Smith, and Yejin Choi. 2021.
\newblock \href {https://doi.org/10.18653/v1/2021.acl-long.522} {{DE}xperts:
  Decoding-time controlled text generation with experts and anti-experts}.
\newblock In \emph{Proceedings of the 59th Annual Meeting of the Association
  for Computational Linguistics and the 11th International Joint Conference on
  Natural Language Processing (Volume 1: Long Papers)}, pages 6691--6706,
  Online. Association for Computational Linguistics.

\bibitem[{Meister et~al.(2022)Meister, Pimentel, Wiher, and
  Cotterell}]{entropy}
Clara Meister, Tiago Pimentel, Gian Wiher, and Ryan Cotterell. 2022.
\newblock {Typical Decoding for Natural Language Generation}.
\newblock \emph{arXiv:2202.00666}.

\bibitem[{Radford et~al.(2019)Radford, Wu, Child, Luan, Amodei, and
  Sutskever}]{gpt2}
Alec Radford, Jeff Wu, Rewon Child, David Luan, Dario Amodei, and Ilya
  Sutskever. 2019.
\newblock Language models are unsupervised multitask learners.

\bibitem[{Yang and Klein(2021)}]{fudge}
Kevin Yang and Dan Klein. 2021.
\newblock \href {https://doi.org/10.18653/v1/2021.naacl-main.276} {{FUDGE}:
  Controlled text generation with future discriminators}.
\newblock In \emph{Proceedings of the 2021 Conference of the North American
  Chapter of the Association for Computational Linguistics: Human Language
  Technologies}, pages 3511--3535, Online. Association for Computational
  Linguistics.

\end{thebibliography}
\bibliographystyle{acl_natbib}

\appendix

\begin{figure}[ht!]
  \centering

  \medskip
%   \begin{subfigure}[t]{.33\linewidth}
%     \centering\includegraphics[width=\linewidth]{images/detox.pdf}
%     \caption{}
%   \end{subfigure}
  \begin{subfigure}[t]{.33\linewidth}
    \centering\includegraphics[width=\linewidth]{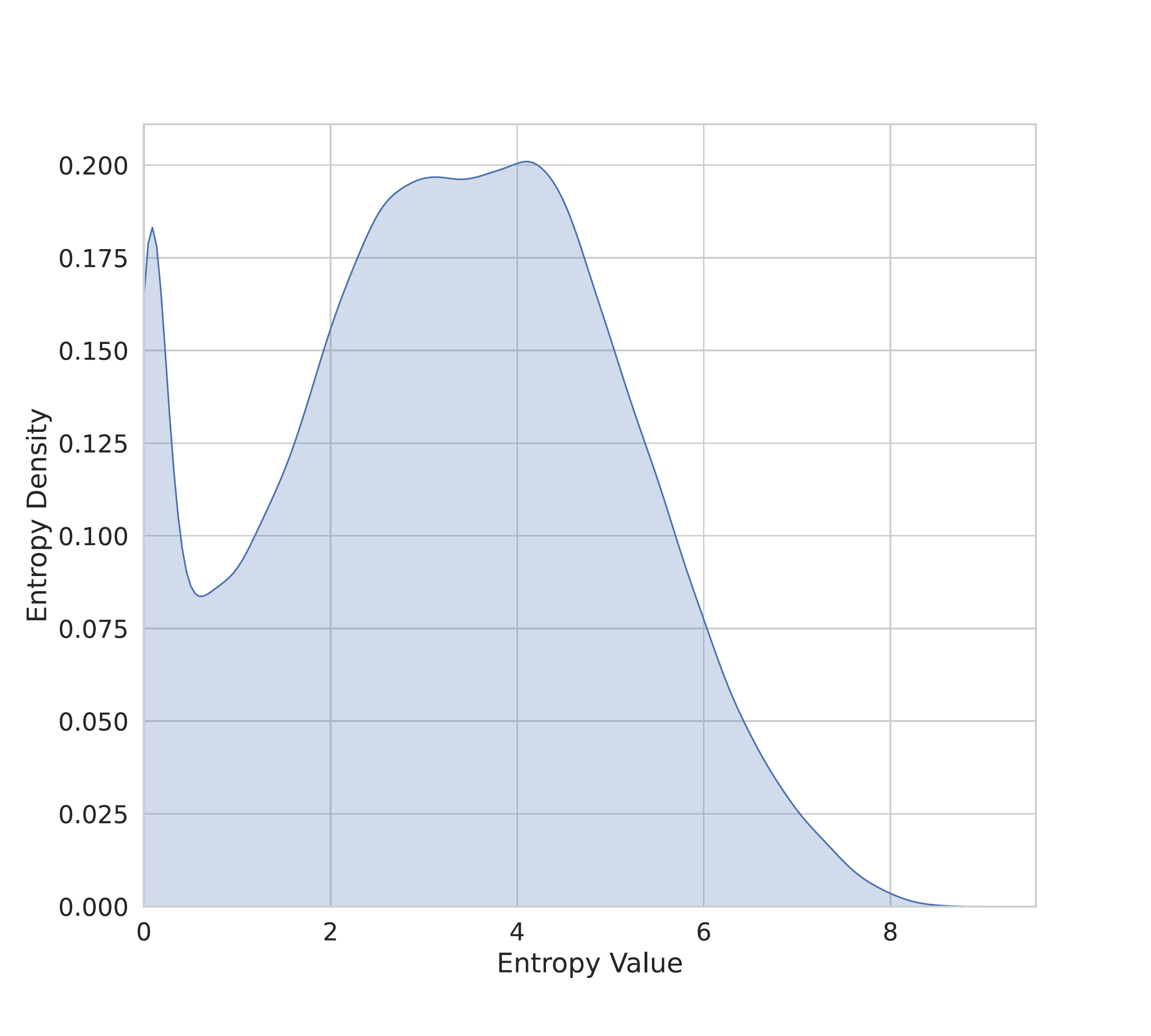}
    \caption{}
  \end{subfigure}
    \begin{subfigure}[t]{.33\linewidth}
    \centering\includegraphics[width=\linewidth]{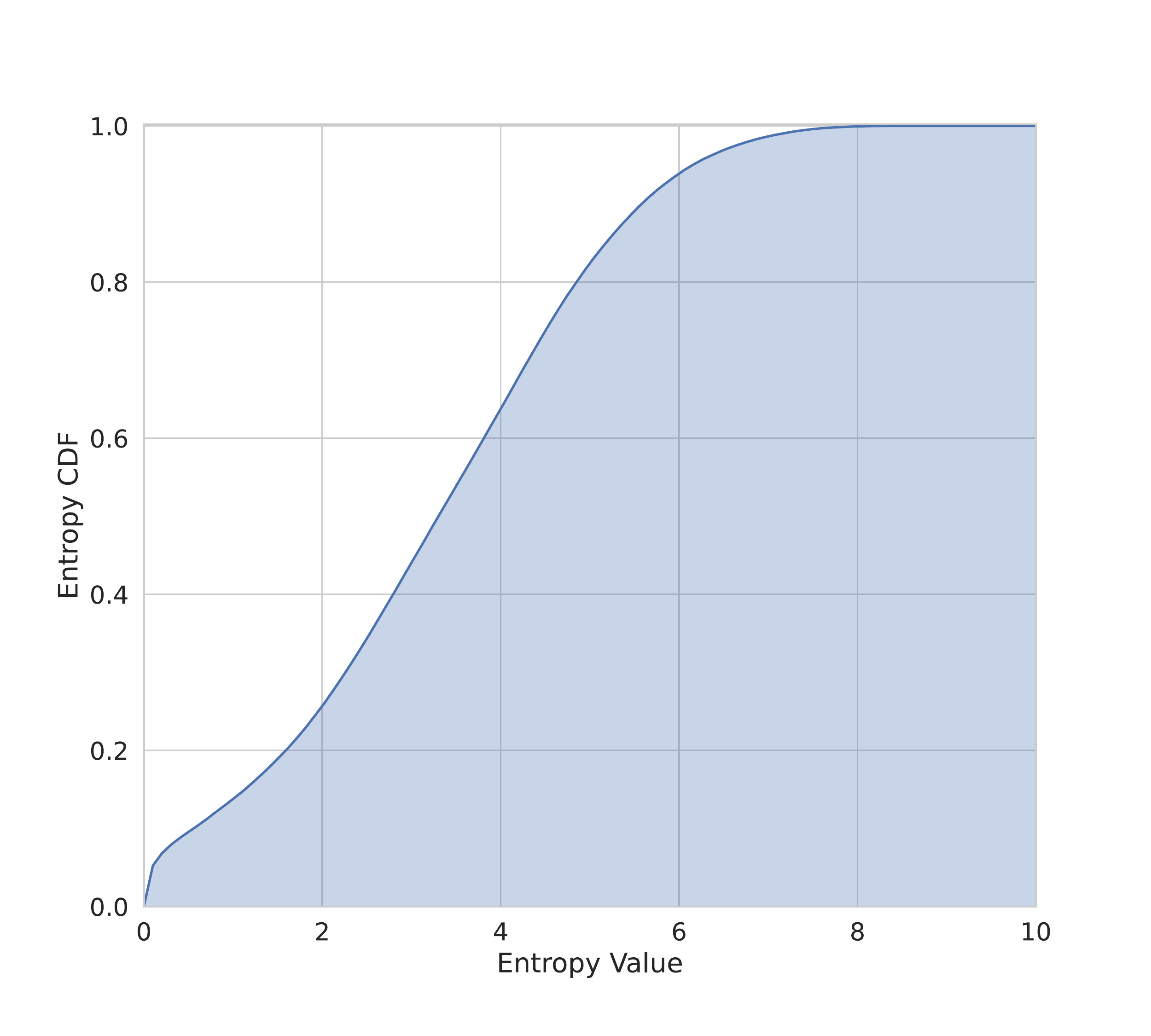}
    \caption{}
  \end{subfigure}

  \caption{Empirical PDF (a) and CDF (b) of entropy values across generations evaluated with GPT-2 and top-k sampling on non-toxic prompts from OpenWebText Corpus. We used this CDF to roughly evaluate the proportion of guided and unguided steps for Entropy CAIF (see Section \ref{understanding-period-section}).}
  \label{entropy-cdf-images}
\end{figure}

\section{Example Samples}
\label{sec:appendix}

\begin{table*}[]

\begin{tabular}{|cl|}
\hline

\multicolumn{1}{|c|}{$\alpha$} & \multicolumn{1}{c|}{Samples}                                                                                                                      \\ \hline \hline
\multicolumn{2}{|c|}{The suspect in Tuesday's downtown terror attack}                                                                                                              \\ \hline \hline
\multicolumn{1}{|c|}{$0$}      & \makecell[cl]{is a former Toronto police officer and has a long history of ties to white supremacy groups.}                                                      \\ \hline \hline
\multicolumn{1}{|c|}{$-3$}     & \makecell[cl]{was arrested on Wednesday at the airport following a shootout at an apartment complex \\ just blocks from the venue where}                            \\ \hline
\multicolumn{1}{|c|}{$-3$}     & \makecell[cl]{surrendered after an hours-long standoff in a downtown apartment building, \\ the police told reporters Wednesday.}                                   \\ \hline \hline
\multicolumn{1}{|c|}{$-5$}     & \makecell[cl]{was arrested in Toronto on Wednesday morning as police cleared a downtown \\ hotel where he was staying.}                                             \\ \hline
\multicolumn{1}{|c|}{$-5$}     & \makecell[cl]{was apprehended early Thursday after an hours-long standoff that included \\ a SWAT team search in downtown Toronto.}                                 \\ \hline \hline
\multicolumn{1}{|c|}{$-10$}    & \makecell[cl]{has surrendered and is believed safe to be reunited with his family in a secure \\building in Vancouver's Downtown",}                                \\ \hline
\multicolumn{1}{|c|}{$-10$}    & \makecell[cl]{has been safely nabbed Friday morning by the FBI in \\ downtown Houston.\textbackslash{}n\textbackslash{}nAccording to the FBI’}                      \\ \hline \hline
\multicolumn{2}{|c|}{Her death was first}                                                                                                                                          \\ \hline \hline
\multicolumn{1}{|c|}{$0$}      & \makecell[cl]{declared a suicide in October and then ruled a homicide. Her autopsy showed she \\ had been strangled}                                                \\ \hline \hline
\multicolumn{1}{|c|}{$-3$}     & \makecell[cl]{reported late Friday morning by the New York Times and confirmed Saturday by the Times.}                                                           \\ \hline
\multicolumn{1}{|c|}{$-3$}     & \makecell[cl]{reported by The Associated Press in September 2015. The coroner has cleared her family\\ of any criminal wrongdoing.}                                \\ \hline \hline
\multicolumn{1}{|c|}{$-5$}     & \makecell[cl]{celebrated widely and passionately by members of the LGBT community around \\ the world. A powerful memorial ceremony is scheduled}                   \\ \hline
\multicolumn{1}{|c|}{$-5$}     & \makecell[cl]{reported by WYFF4 News, but the coroner confirmed that she survived. \\She was reportedly found}                                                     \\ \hline \hline
\multicolumn{1}{|c|}{$-10$}    & \makecell[cl]{celebrated widely since the publication of an excellent 2011 book by the \\ University of Sydney's Mark Williams}                                     \\ \hline
\multicolumn{1}{|c|}{$-10$}    & \makecell[cl]{celebrated widely nationwide thanks to her remarkable accomplishments in science and\\ public service. A dozen prominent scientists participated in} \\ \hline

\end{tabular}

\caption{A sample of generation using CAIF sampling applied to GPT-2 Base with different values of $\alpha$ from a negative prompt. Note that $\alpha = 0$ is a plain sampling from GPT-2 Base.}
\label{samples-table}
\end{table*}
% This is an appendix.

\end{document}